\documentclass[a4paper,fleqn]{cas-dc}

\usepackage[numbers]{natbib}

\usepackage{microtype}
\usepackage{algorithm}
\usepackage{algorithmic}
\usepackage{listings}
\usepackage{xcolor}
\usepackage{amsmath}
\usepackage{amsthm}
\usepackage{tabularx}
\usepackage{booktabs}
\usepackage{chngcntr}
\usepackage{float}
\usepackage{enumitem}

\newcommand{\mcode}[1]{\(\mathtt{#1}\)}

\newtheoremstyle{break}
  {\topsep}   
  {\topsep}   
  {\itshape}  
  {}          
  {\bfseries} 
  {:}         
  {\newline}  
  {}          
\theoremstyle{break}

\definecolor{codegreen}{rgb}{0,0.6,0}
\definecolor{codegray}{rgb}{0.5,0.5,0.5}
\definecolor{codepurple}{rgb}{0.58,0,0.82}
\definecolor{backcolour}{rgb}{0.95,0.95,0.92}

\lstdefinestyle{mystyle}{
    commentstyle= \color{red!50!green!50!blue!50},
    keywordstyle= \color{blue!70},
    numberstyle=\tiny\color{codegray},
    stringstyle=\color{codepurple},
    basicstyle=\ttfamily\footnotesize,
    breakatwhitespace=false,
    breaklines=true,
    captionpos=b,
    keepspaces=true,
    numbers=left,
    numbersep=5pt,
    showspaces=false,
    showstringspaces=false,
    showtabs=false,
    tabsize=2,
    frame=single  
}

\lstdefinelanguage{md}{
  morekeywords={[1]\#,\#\#,\#\#\#,\*,\*\*,\_,\-\ ,\>,\`},
  sensitive=false,
  morecomment=[l]{\%},   
  morestring=[b]",      
  alsoletter={\#\`*\-},
}



\begin{document}
\let\WriteBookmarks\relax
\def\floatpagepagefraction{1}
\def\textpagefraction{.001}
\shorttitle{}
\shortauthors{Fan et~al.}

\title [mode = title]{
CADDesigner: Conceptual CAD Model Generation with a General-Purpose Agent
}      

\author[]{Fengxiao Fan}
\cormark[1]

\author[]{Jingzhe Ni}
\cormark[1]

\author[]{Xiaolong Yin}

\author[]{Sirui Wang}

\author[]{Xingyu Lu}

\author[]{Qiang Zou}

\author[]{Ruofeng Tong}

\author[]{Min Tang}

\author[]{Peng Du}
\cormark[2]

\affiliation{
  organization={Zhejiang University},
  country={China}
}

\cortext[cor1]{Equal contributions.}
\cortext[cor2]{Corresponding author: dp@zju.edu.cn}

\begin{abstract}
Computer-Aided Design (CAD) is widely used for conceptual design and parametric 3D modeling, but typically requires a high level of expertise from designers. To lower the entry barrier and facilitate early-stage CAD modeling, we present CADDesigner, an LLM-powered agent for conceptual CAD design. The agent accepts both textual descriptions and sketches as input, engaging in interactive dialogue with users to refine and clarify design requirements through comprehensive requirement analysis. Built upon a novel Explicit Context Imperative Paradigm (ECIP), the agent generates high-quality CAD modeling code. During the generation process, the agent incorporates iterative visual feedback to improve model quality. Generated design cases can be stored in a structured knowledge base, providing a mechanism for continual knowledge accumulation and future improvement of code generation. Experimental results show that CADDesigner achieves competitive performance and outperforms representative baselines on conceptual CAD model generation tasks.
\end{abstract}




\maketitle

\section{Introduction}
Conceptual design of Computer-Aided Design (CAD) models often begins with incomplete or abstract specifications, requiring designers to translate rough ideas into precise parametric models. Creating such 3D models using traditional CAD software typically demands substantial expertise and familiarity with complex modeling operations, which can be a barrier for novice users and slow down early-stage product development. Traditional CAD platforms--such as OnShape, AutoCAD, SolidWorks, and CATIA--primarily rely on manual sketching and extrusion workflows, making iterative prototyping time-consuming and error-prone. However, recent advances in artificial intelligence, particularly the emergence of LLMs, present promising opportunities to automate CAD model generation, thereby reducing the entry barrier and accelerating the design workflow.

Early research on automatic CAD model generation has primarily focused on parametric modeling approaches~\cite{wu2021deepcad,li2025mamba,khan2024cad,text2cad}. However, these methods are constrained by the limited diversity of training data and the representational capacity of their model outputs, supporting only a narrow set of CAD operations. Recent work leveraging LLMs for CAD code generation has shown promise in overcoming these limitations~\cite{li2025cad,wang2025cad,rukhovich2024cad}. Fine-tuned LLMs can interpret multimodal user inputs and generate corresponding CAD modeling code. Nevertheless, fine-tuning open-source LLMs requires substantial GPU resources, and high-quality CAD training datasets remain scarce, limiting the diversity and quality of the generated code.

To address these challenges, we propose CADDesigner, an LLM-powered agent for conceptual CAD design. CADDesigner accepts both textual descriptions and sketches as input and engages in interactive dialogue with users to refine design requirements. The system combines knowledge-constrained code generation with iterative visual feedback, enabling the agent to generate high-quality CAD modeling code that better aligns with user design intent.

\begin{figure*}[!ht]
  \centering
  \includegraphics[width=1.0\textwidth]{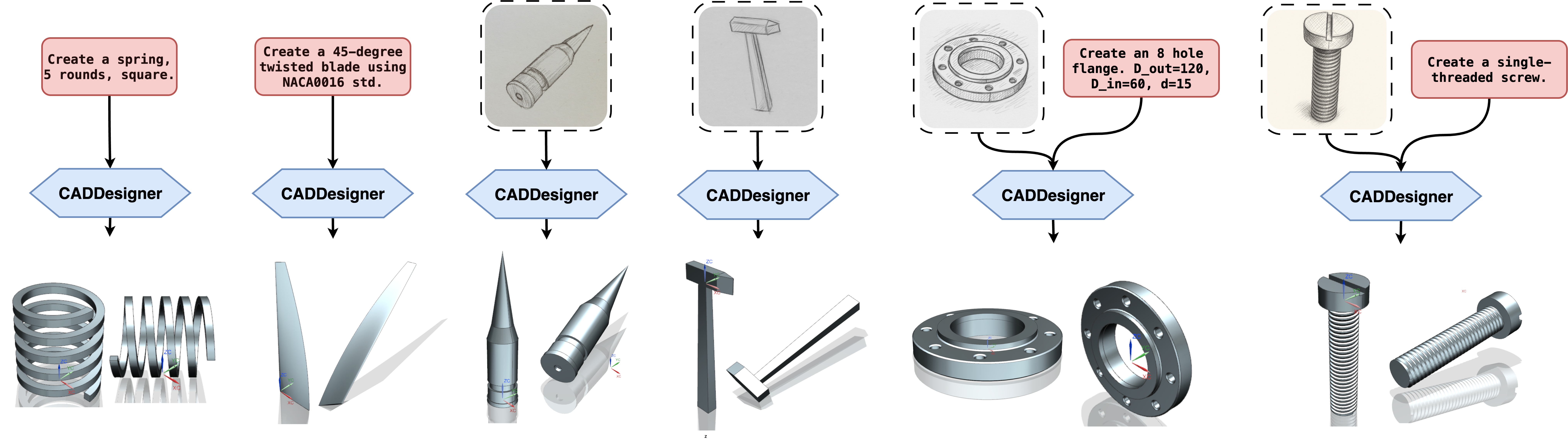} 
  \caption{Demonstration of various CAD models generated by CADDesigner. Our method supports multimodal input and a broad range of CAD operations, including extrusion, revolution, fillet/chamfer, sweeping, lofting, etc., as well as the creation of standard components such as flanges and screws.
  }
  \label{fig:show-muscle}
\end{figure*}

To further improve code generation quality, we propose the Explicit Context Imperative Paradigm (ECIP), an LLM-oriented representation style implemented through a lightweight API layer built on top of CadQuery~\cite{cadquery}. ECIP explicitly represents modeling context and restructures CAD operations into an explicit imperative workflow, enabling clearer state transitions, more reliable code generation, and easier iterative correction while remaining aligned with CadQuery semantics. Representative results generated by CADDesigner are shown in Figure~\ref{fig:show-muscle}, and our main contributions are summarized as follows:

\begin{itemize}[leftmargin=*]
    \item We present CADDesigner, a framework for conceptual CAD model generation. CADDesigner accepts textual descriptions and sketches as user input, and integrates tools for requirement analysis, knowledge-constrained code generation, and vision-based error correction to generate CAD scripts that satisfy user intent.
    \item 
    We introduce an explicit context imperative paradigm for CAD modeling script generation based on an imperative function-calling structure.
    This paradigm decouples individual CAD operations and improves code generation quality through explicit type annotations, LLM-friendly error representations, and self-evolving capabilities. It supports a wide range of operations, including extrusion, revolution, fillet, chamfer, sweeping, lofting, etc.
\end{itemize}

The remainder of this paper is organized as follows. We begin with a review of related work. We then introduce our proposed CAD modeling code generation framework, CADDesigner, along with the novel code generation paradigm designed to enhance code quality. Next, we describe our experimental setup and present the results. Finally, we provide a comprehensive analysis and comparison to highlight the strengths and limitations of the proposed approach.

\section{Related Work}
This section provides a comprehensive review of four key areas relevant to our work: parametric CAD modeling, LLM-driven CAD generation, agent-based model generation, and parametric CAD modeling SDKs.

\subsection{Parametric CAD Model Generation}

Parametric CAD model generation approaches commonly employ supervised learning techniques to produce sequences of CAD commands, which can be imported into CAD modeling software to generate editable, parametric files~\cite{willis2020fusion, wu2021deepcad}. Systems such as DeepCAD~\cite{wu2021deepcad}, Fusion 360 Gallery~\cite{willis2020fusion}, SkexGen~\cite{xu2022skexgen}, and Diffusion-CAD~\cite{zhang2025diffusion} adopt customized neural network architectures trained to generate command sequences. However, they primarily support only basic modeling operations such as sketching and extrusion, and struggle to generalize to more complex modeling procedures.

Additional efforts~\cite{khan2024cad, guo2022complexgen, ma2024draw} explore CAD model reconstruction from point cloud data, aiming to recover structured geometry from unorganized 3D observations. More recent methods~\cite{li2025caddreamer, chen2025cadcrafter} further attempt to infer structured, editable CAD models directly from unstructured RGB images (either single- or multi-view), bridging the gap between 2D visual understanding and 3D CAD modeling. Despite these advancements, existing approaches remain limited in their ability to generate complex CAD models, primarily due to constrained training data diversity and limited output representation capacity.

\subsection{LLM-based CAD Model Generation}

Fine-tuned LLMs and vision-language models (VLMs) have emerged as a prominent direction for CAD model generation, enabling the synthesis of CAD command sequences or Python code from multimodal input. Cad-LLM~\cite{wu2023cad} and CadVLM~\cite{wu2024cadvlm} utilize fine-tuned LLMs to generate parametric models from sketches and images. CAD-MLLM~\cite{xu2024cad} targets multimodal inputs--including text, images, and point clouds--for parametric model generation. CAD-Llama~\cite{li2025cad} proposes a hierarchical annotation pipeline that transforms command sequences into semantically structured CAD code. This is followed by adaptive pre-training and instruction tuning, enabling LLMs to generate high-fidelity parametric 3D models. CAD-Recode~\cite{rukhovich2024cad} converts point clouds into sequential representations, then uses a pre-trained LLM (Qwen2-1.5B~\cite{team2024qwen2}) to generate corresponding CAD code. Text-to-CadQuery~\cite{xie2025text} directly maps text descriptions to CadQuery code, leveraging pre-trained LLMs' capabilities in Python generation and spatial reasoning.

While these methods demonstrate strong generation capabilities, they remain limited by several factors: the computational cost of fine-tuning large models, the closed nature of many commercial LLMs, and the scarcity of high-quality, diverse training datasets for CAD.

\subsection{Agent-based CAD Model Generation}

Another research direction focuses on constructing CAD generation agents using prompt engineering and ultra-large foundation models, circumventing the need for additional training. CAD-Assistant~\cite{mallis2024cad} presents a general-purpose CAD agent framework that integrates visual and language models (VLLMs) as planners to generate Python code for FreeCAD~\cite{freecad}, achieving zero-shot capability across diverse CAD tasks. SeekCAD~\cite{li2025seek} combines retrieval-augmented generation (RAG), visual feedback, and a chain-of-thought (CoT) mechanism to generate customized CAD modeling code with self-optimization.
3D-PreMise~\cite{yuan20243d} investigates the potential and limitations of LLMs in program synthesis for manipulating 3D software and generating parametrically controlled shapes. CADCodeVerify~\cite{alrashedy2024generating} incorporates iterative validation and refinement loops, using visual language models to pose and answer verification questions about the generated CAD outputs.

In contrast to these works, CADDesigner accepts multimodal user inputs--text and sketches--to capture detailed design requirements, engages in interactive refinement with users, and enhances code quality and model fidelity through visual feedback and iterative optimization.

\begin{figure*}[!ht]
  \centering
  \includegraphics[width=0.96\textwidth]{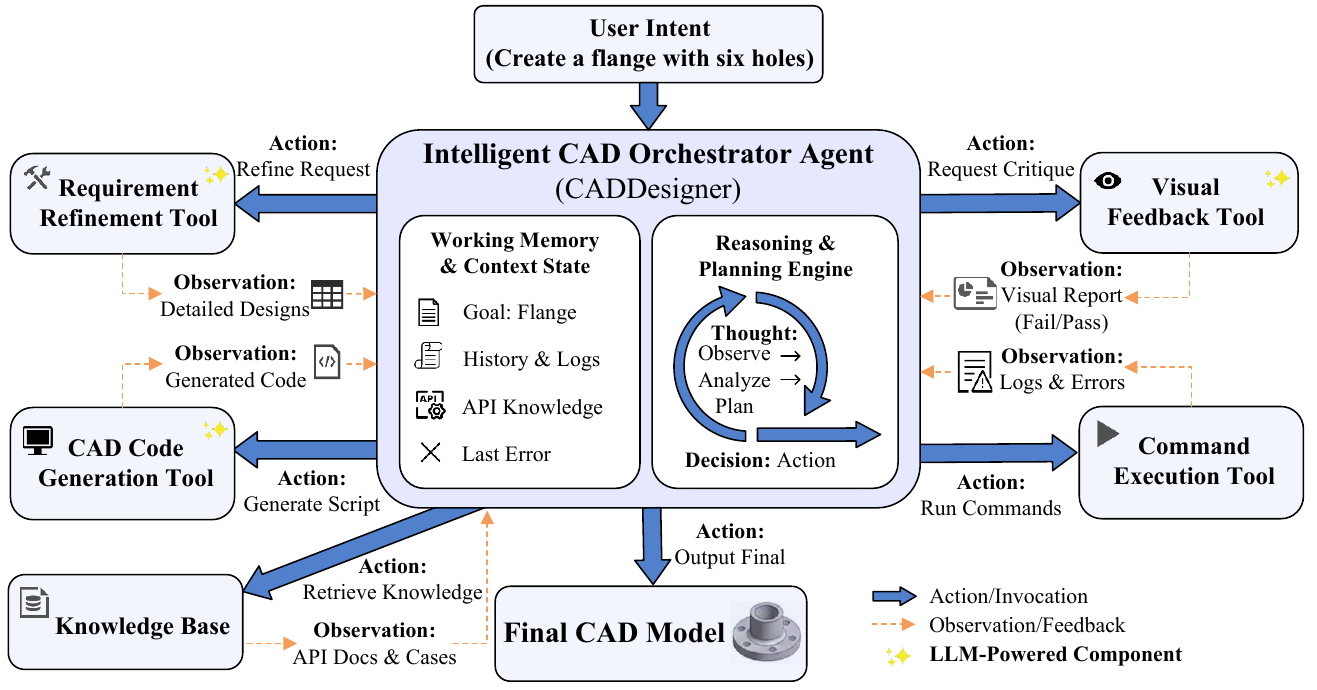} 
  \caption{The Intelligent CAD Orchestrator Agent, CADDesigner, follows a ReAct-style paradigm to progressively transform user requirements into valid CAD models through iterative reasoning, tool execution, and feedback refinement. It first refines user requirements into detailed designs, generates executable modeling code using domain APIs, and analyzes execution results via both symbolic (e.g., shell logs and errors) and visual feedback (e.g., rendered 3D views).}
  \label{fig:pipeline}
\end{figure*}

\subsection{Parametric CAD Modeling SDKs}

Beyond modeling methods, existing CAD modeling SDKs play a crucial role as the underlying representations for code-based CAD generation. Most Python-based parametric CAD frameworks are built upon OpenCASCADE Technology (OCCT), a widely used boundary representation modeling kernel. CadQuery provides a high-level Python interface with a Fluent API based on method chaining and workplane abstractions~\cite{cadquery}, enabling concise modeling workflows through implicit context propagation. Build123d adopts a different design based on context managers (e.g., $\mathtt{with}$ statements) and operator-style modeling by overloading operators in Python, offering tighter integration with native Python constructs and improved modularity~\cite{build123d}.

However, these SDKs are not specifically designed for LLM-based code generation. CadQuery's Fluent API relies on implicit internal state transitions, which can obscure intermediate modeling context and increase the difficulty for LLMs to track dependencies across operations. Despite its explicit context declaration, build123d's reliance on symbolic operators ($\mathtt{@}$, $\mathtt{+}$, $\mathtt{-}$) for complex modeling may make semantic intent less transparent for LLM-based code generation, especially compared with explicit function-call interfaces.
Such implicit context handling and lack of well-defined semantics can lead to ambiguity and reduced reliability in generated code. These limitations motivate the need for a more explicit and LLM-friendly CAD representation, which we address through the proposed Explicit Context Imperative Paradigm (ECIP).

\section{CAD Modeling Architecture}

In this section, we introduce the CADDesigner architecture, which features a ReAct-style agent interacting with a diverse set of integrated CAD tools to iteratively generate and refine CAD models. The overall framework is illustrated in Figure~\ref{fig:pipeline}.

\subsection{ReAct Agent rather than Workflow}
CADDesigner employs a ReAct-style~\cite{yao2023react} agent-centric architecture, where a central agent governs the entire modeling loop through iterative reasoning, tool execution, and feedback integration.

Unlike traditional hard-coded workflows, our architecture leverages prompt-based workflow specifications, where the system prompts encode procedural guidance for the agent. This design allows for flexible and revisable task orchestration, as the workflow is not statically programmed but interpreted at runtime by the agent. As a result, users can intervene at any stage of the process, like adding constraints, correcting directions, or adjusting goals, leading to a truly interruptible and human-in-the-loop modeling cycle.

The ReAct-style loop includes reasoning, acting, and reflecting: the agent expands coarse inputs into parameterized specs, generates executable code, renders outputs, evaluates results, and iteratively refines the model until user intent is satisfied. By consolidating control within the agent, we enable adaptive tool use and interpretable, interactive CAD automation in under-specified or evolving design tasks.

\subsection{CAD Tools}

The intelligent orchestration relies on a diverse suite of CAD-oriented tools. Each tool addresses a specific stage within the design-to-modeling pipeline and works in synergy under the direction of the ReAct agent. This subsection describes the CAD tools integrated in our pipeline.

\subsubsection{Core CAD Toolset}

CADDesigner employs various CAD-specific tools to support its design-to-modeling pipeline. Central to this set are four tools, denoted $\mathcal{T} = \{T_1, T_2, T_3, T_4\}$, each fulfilling a distinct role.
\begin{itemize}[leftmargin=*]
    \item \textbf{Requirement Refinement Tool $T_1 \colon \mathcal{I} \to \mathcal{D}$} maps user-provided multimodal input $\mathcal{I}$ to detailed design descriptions $\mathcal{D}$, refining initial requirements into structured design specifications.
    \item \textbf{Code Generation Tool $T_2 \colon \mathcal{D} \to \mathcal{C}$} translates finalized design descriptions into executable CAD modeling code $\mathcal{C}$. A new code paradigm for CAD code generation has been designed and used by $T_2$. This is discussed in detail in Section~\ref{sec:cad_code_paradigm}.
    \item \textbf{Command Execution Tool \(T_3 \colon \mathcal{C} \to \mathcal{M}\)} executes the generated Python code \(\mathcal{C}\) to produce the corresponding CAD model \(\mathcal{M}\) by running shell commands.
    \item \textbf{Visual Feedback Tool $T_4 \colon \{\mathcal{M}, \mathcal{D}\} \to (P, F)$}
    outputs both a high-level termination signal $P \in \{0, 1\}$, where $P = 1$ triggers termination, and structured diagnostic feedback $F$ for refinement.
    Inside this tool, a multi-view rendering result of model $\mathcal{M}$ will be generated. We denote this render result as $\mathcal{V}$. Thus $T_4$ can be separated into 2 steps: $T_4^1: \mathcal{M} \to \mathcal{V}$, and $T_4^2: \{\mathcal{V}, \mathcal{D}\} \to (P, F)$
\end{itemize}

The sequential composition of these core tools forms the main functional pipeline:
\begin{equation}
T_4^1 \circ T_3 \circ T_2 \circ T_1: \mathcal{I} \to \mathcal{V}.
\end{equation}

This pipeline is augmented with a closed-loop correction mechanism: after generating visual feedback $\mathcal{V}$, the system invokes $T_4^2$ to obtain $(P, F)$. If $P = 0$, the feedback $F$ is used to guide revised design generation via $T_3 \circ T_2 \circ T_1$. If $P = 1$, the process terminates successfully. Otherwise, the loop continues until $T_4^2$ returns a termination signal.

\subsubsection{Auxiliary Toolset}

In addition to the core CAD toolset, CADDesigner integrates several auxiliary tools that support efficient context management and system interaction. These include:

\begin{itemize}[leftmargin=*]
    \item \textbf{SketchPad Tool}: A flexible key-value storage tool designed to mitigate the context explosion problem in LLM-based agents. SketchPad allows the agent to store arbitrary intermediate data at any time during the design process, including image paths, reference code snippets, execution results, and more. By storing context data as keyed entries instead of embedding entire data repeatedly, it significantly reduces LLM context size. SketchPad also supports automatic summarization, tag-based search, and automatic expiration of outdated entries, enabling flexible and scalable context management.

    \item \textbf{File Operation Tools}: These tools provide fundamental file-reading and file-writing capabilities, allowing the agent to store or load files as required throughout the CAD design pipeline. Similar to SketchPad, file content or references are stored and accessed efficiently to avoid large context overheads.
\end{itemize}

Together, these auxiliary tools complement the core CAD toolset by enabling robust, efficient, and scalable operation of the CADDesigner agent in complex multi-turn interactions.

\subsection{Requirements Analysis}

During the conceptual design phase, designers typically have only rough descriptions or sketches, making it challenging to precisely define detailed model parameters as well as the modeling process. To address this, we support designers by conducting requirement analysis to clarify and refine their design intention.

Let the set of initial information provided by the user be \(\mathcal{I} \subseteq \mathcal{I}_{\mathrm{txt}} \times \mathcal{I}_{\mathrm{img}}\), where \(\mathcal{I}_{\mathrm{txt}}\) denotes the set of possible text inputs (including empty text \(\emptyset\)) and \(\mathcal{I}_{\mathrm{img}}\) denotes the set of possible image inputs (including a null image \(\emptyset\)). Let \(\mathcal{P}\) be a set of parameters sufficient for generating a CAD model for the current design task. The process by which CADDesigner conducts detailed design can be expressed as \(\mathcal{F}_{\text{design}}: \mathcal{I} \to \mathcal{D}_{\text{detail}}\), where \(\mathcal{D}_{\text{detail}}\) denotes the detailed design containing specific parameters, satisfying \(\mathcal{P} \subseteq \mathcal{D}_{\text{detail}}\), and this process is accomplished by \(T_1\). At this stage, CADDesigner enables users from different disciplines to realize their design intentions through interaction. The user's revision process for the detailed design is modeled as an iterative function:
\begin{equation}
    R: \mathcal{D}_{\text{detail}} \times \mathcal{U} \to \mathcal{D}_{\text{detail}},
\end{equation}
where \(\mathcal{U}\) is the set of user revision operations. After \(n\) iterations of revision, we obtain \(\mathcal{D}_{\text{final}} = R^n(\mathcal{D}_{\text{detail}}, \mathcal{U})\), which is then passed to the next stage.

\subsection{Knowledge Constrained Code Generation}\label{sec:codegen}
\(\mathcal{D}_{\text{final}}\) is the detailed design confirmed by the user. The process by which the code generation tool \(T_2\) receives \(\mathcal{D}_{\text{final}}\) and generates CAD modeling code is expressed as \(G: \mathcal{D}_{\text{final}} \to \mathcal{C}\).
A structured knowledge base \(\mathcal{K}\) is constructed to support code generation. \(\mathcal{K}\) includes a set of function annotations \(Anno \subseteq \mathcal{K}\) and a set of basic model cases \(Case \subseteq \mathcal{K}\), which provide semantic and structural references for the code generation tool. With the assistance of the knowledge base, the code generation process is extended to:
\begin{equation}
    G': \mathcal{D}_{\text{final}} \times \mathcal{K} \to \mathcal{C}',
\end{equation}
where \(\mathcal{C}'\) denotes the code generated by referencing the knowledge base, which is semantically and structurally richer compared to the code \(\mathcal{C}\) generated without using the knowledge base. The execution result of the code, including detailed geometry metadata (see Section~\ref{sec:core_module}, Metadata \& Tagging) such as volumes and measurements of some parts of the model, is returned after execution. If the execution succeeds, the code can run successfully and generate a model \(\mathcal{M} = T_3(\mathcal{C}')\), which is then passed to the next stage; if the execution fails, structured error information is returned to guide the next iteration of code generation \(G'\) to produce new code. This feedback loop helps to iteratively correct the code and improve the success rate. Therefore, the knowledge base provides essential semantic and structural support for code generation, and the iterative process combined with execution feedback further enhances the likelihood of successful code generation. In addition, execution-time geometric metadata and queryable CAD-native representations provide structured signals for dimension-sensitive and constraint-related verification beyond visual inspection.

\subsection{Vision-based Iterative Error Correction}
Multi-view snapshots of the generated CAD model \(\mathcal{M}\) are taken by \(T_4^1\), producing an image set \(\mathcal{V} = \{ \text{Img}_1,\allowbreak \text{Img}_2,\allowbreak \dots,\allowbreak \text{Img}_6 \}\) (corresponding to the Front-Top-Left, Back-Bottom-Right, Back-Top-Left, Front-Bottom-Right, Top, and Right views respectively).
CADDesigner evaluates whether the model meets the user's intent \(\mathcal{D}_{\text{final}}\) by using both a high-level termination signal \(P \in \{0,1\}\) and structured diagnostic feedback generated from multi-view observations.

Internally, \(P\) is determined by \(T_4^2\), which involves two components:
\begin{itemize}[leftmargin=*]
    \item \textbf{Visual Question Generation} \(F_{\mathrm{vq}}: \mathcal{V} \times \mathcal{D}_{\text{final}} \to \mathcal{Q}\), which produces a set of targeted visual questions \(\mathcal{Q} = \{q_1, q_2, \ldots, q_n\}\) based on the user's requirements and the multi-view images \(\mathcal{V}\).
    \item \textbf{Visual Feedback Generation} \(F_{\mathrm{vf}}: \mathcal{V} \times \mathcal{Q} \to (P, F)\), which analyzes the multi-view images \(\mathcal{V}\) to answer the questions in \(\mathcal{Q}\), and outputs explicit visual feedback \(F\) along with the binary decision \(P\). When \(P = 0\), it returns visual feedback \(F\) to the code generation stage to regenerate the code; when \(P = 1\), the result is delivered to the user for final judgment. The user's satisfaction with the modeling result is determined by \(Sat: \mathcal{M} \to \{0,1\}\). If \(Sat(\mathcal{M}) = 1\), \(\mathcal{M}\) is added to the case library \(\mathcal{K}\), that is, \(Case' = Case \cup \{\mathcal{M}\}\); if \(Sat(\mathcal{M}) = 0\), it re-enters the code generation stage.
\end{itemize}

In the entire process, except for the user feedback links (\(R\) and \(Sat\)), all state transitions \(Trans: State \to State'\) are determined independently by CADDesigner.

\section{CAD Code Generation Paradigm}
\label{sec:cad_code_paradigm}

In this section, we provide a detailed description of a new CAD code generation paradigm from four aspects: the Explicit Context Imperative Paradigm (ECIP), its practical realization via an ECIP-based CAD API, LLM-friendly CAD code design, and the construction of a reusable modeling knowledge base.

\subsection{Explicit Context Imperative Paradigm}
\label{sec:ecip}

In recent years, many works have employed \textbf{CadQuery} (CQ) code as an intermediate representation for CAD model generation. CadQuery commonly employs a Fluent API design paradigm~\cite{cadquery}, where operations are expressed via chained method calls to enable a highly readable and compact modeling workflow. This style relies on an \emph{implicit internal context} \(C\) that tracks the current modeling state and is automatically passed between operations. Formally, each operation \(f_k\) updates the hidden state \(C_k\) with parameters \(p_k\):
\vspace{-8pt}
\begin{equation}
C_{k+1} = f_k(C_k, p_k).
\end{equation}
A chained expression can therefore be written as:
\begin{equation}
C_n = f_n \circ f_{n-1} \circ \cdots \circ f_1 (C_0).
\end{equation}

While concise, this implicit state can make semantic dependencies less explicit, which may complicate automated code generation for LLMs that must track intermediate contexts. In contrast, low-level APIs like Python\-OCC require explicit inputs for each operation, yielding clear semantics but verbose, fine-grained code that lacks high-level abstraction and usability. 

To address these limitations, we introduce the \textbf{Explicit Context Imperative Paradigm (ECIP)}, which enforces explicit passing of all input objects and parameters to each operation. Formally, each operation \(g_k\) maps explicit input state \(S_k\) and parameters \(p_k\) to a new state \(S_{k+1}\):
\begin{equation}
S_{k+1} = g_k(S_k, p_k).
\end{equation}

Thus, the modeling workflow is represented as a sequence of explicit state transformations:
\begin{equation}
S_1\!=\!g_1(S_0,p_1),\,S_2\!=\!g_2(S_1,p_2),\,\ldots,\,S_n\!=\!g_n(S_{n-1},p_n),
\end{equation}
where all intermediate states \(S_k\) are explicit variables. 

\begin{figure}[tb]
    \centering
    \includegraphics[width=0.48\textwidth]{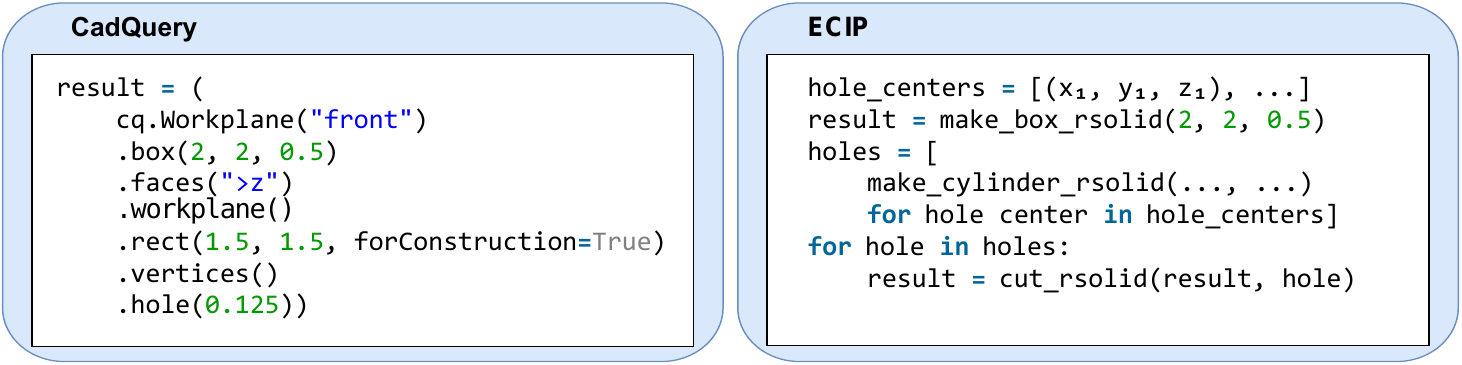}
    \vspace{-4pt}
    \caption{Code comparison between CadQuery (left) and ECIP (right). ECIP explicitly passes context and supports standard Python constructs, improving code clarity and flexibility.}
    \label{fig:cq_vs_ecip}
\end{figure}

Figure~\ref{fig:cq_vs_ecip} illustrates a typical code comparison between CadQuery and ECIP. CadQuery (left) uses chained method calls that implicitly pass context, making the code compact but less flexible for inserting standard Python statements or loops. In contrast, ECIP (right) explicitly passes all inputs, allowing the use of standard Python control flow like \(\mathtt{for}\) loops and variable assignments, which enhances modularity, readability, and ease of debugging. Thus, ECIP combines the clarity of low-level APIs with the expressiveness of high-level modeling, improving the accuracy and interpretability of agent-driven CAD code generation.
This does not imply that CadQuery cannot support more explicit coding styles; rather, ECIP standardizes such explicitness in a form that is more suitable for LLM-oriented generation and refinement.

\subsection{ECIP-Compliant CAD API Design}

ECIP (referred to as SimpleCADAPI in the actual project) is designed as a command-style Python API built on top of CadQuery's $\mathtt{occ\_impl.shapes}$ module. It serves as an LLM-oriented intermediate representation that remains semantically aligned with CadQuery. The geometric classes (e.g., $\mathtt{Vertex}$, $\mathtt{Edge}$, $\mathtt{Solid}$) are implemented as lightweight wrappers over CadQuery objects, preserving full interoperability with native CadQuery operations.

The design follows the Open-Closed Principle: core geometric types are closed for modification, while operations and extensions are introduced through additional functions. Moreover, SimpleCADAPI is fully compatible with the NumPy library, enabling efficient mathematical computation and geometric transformation directly on CAD entities. It provides explicit input-output interfaces for each operation, facilitating LLM-driven code generation, iterative correction, and modular composition, while retaining the high-level modeling capabilities of CadQuery. SimpleCADAPI is organized into two main parts: the \textbf{Core} module and the \textbf{API} module.

\subsubsection{Core Module}
\label{sec:core_module}
The Core module encapsulates fundamental geometric types and primitive operations. By wrapping the underlying OpenCASCADE \(\mathtt{TopoDS\_Shape}\) classes, it establishes a clear hierarchy and semantic structure for geometric objects, while also managing coordinate systems, supporting tagging and metadata attachment, and enforcing strict type annotations and exception handling.

\begin{itemize}[leftmargin=*]
    \item \textbf{Geometric Object Hierarchy:} The Core module defines a layered geometric object model comprising the following primary classes, all inheriting from \(\mathtt{TaggedMixin}\) to provide unified tagging and metadata management:
    \begin{enumerate}[label=\textbf{(\arabic*)}]
        \item \textbf{Vertex}: 0-dimensional points in 3D space.
        \item \textbf{Edge}: 1-dimensional curves connecting vertices.
        \item \textbf{Wire}: Ordered collections of edges forming open or closed chains.
        \item \textbf{Face}: 2-dimensional bounded surfaces defined by wires (outer boundaries and holes).
        \item \textbf{Shell}: Collections of faces forming partial or complete shells.
        \item \textbf{Solid}: Closed 3-dimensional volumes with definable volumetric properties.
        \item \textbf{Compound}: Arbitrary aggregates of geometric entities for grouping or assembly construction.
    \end{enumerate}

    \item \textbf{Coordinate Systems and Working Planes:} ECIP employs a unified coordinate system framework to ensure consistency and precision in geometric modeling and transformation. This framework includes:
    \begin{enumerate}[label=\textbf{(\arabic*)}]
        \item \textbf{CoordinateSystem}: Represents a 3D spatial frame with an explicit origin, X-axis direction, and normal vector, providing an abstraction for spatial transformations.
        \item \textbf{WORLD\_CS}: A global singleton coordinate system instance defining the default reference frame. It follows the right-hand rule with origin at \([0,0,0]\), the X-axis pointing along the default width direction, the Y-axis along the default height direction, and the Z-axis upward.
        \item \textbf{SimpleWorkplane}: Manages local coordinate systems via context managers, enabling nested coordinate frames with defined origins and orientations. The returned workplane object follows the $\mathtt{\_wp}$ naming convention when the variable is not directly referenced within the block, ensuring linter and LLM compatibility. This simplifies relative geometric operations and supports complex modeling workflows.
    \end{enumerate}

    \item \textbf{Metadata \& Tagging:} ECIP unifies semantic tagging and geometric metadata through a lightweight tag--metadata layer. Each entity maintains user-visible tags and structured metadata, while runtime lineage is stored separately to avoid polluting modeling data. This design supports three complementary mechanisms. First, primitive constructors perform geometry-driven tagging. For example, $\mathtt{make\_box\_rsolid}$ automatically assigns face-level semantic tags such as ``top'', ``bottom'', ``left/right'', and ``side'' based on face normals, edge structure, and primitive type. Second, derived operations attach semantic tags via topology tracking after geometric or topological edits. ECIP wraps OpenCASCADE builders and queries $\mathtt{Modified()}$, $\mathtt{Generated()}$, and $\mathtt{IsDeleted()}$ to compute a $\mathtt{TopoDelta}$. Resulting faces are then automatically assigned operation-aware semantic tags (e.g., $\mathtt{origin.tool}$, $\mathtt{op.cut.generated}$, $\mathtt{origin.body}$). This approach avoids brittle index-based bookkeeping and preserves semantic lineage across boolean and feature operations. Third, ECIP stores structured geometric metadata alongside shapes, including primitive type, box size $(x,y,z)$, cylinder axis and height, sphere center and radius, coordinate system snapshots, topological references ($\mathtt{topo\_ref}$), and compact selector hints (e.g., center, normal, area, length, volume, and bounding box). These metadata are serializable and directly consumable by QL predicates such as $\mathtt{ql.tag(...)}$ and $\mathtt{ql.meta(...)}$.

    \item \textbf{Query Language (QL) for Geometric Selection:} SimpleCADAPI provides a dedicated query language module ($\mathtt{ql}$) for declarative geometric selection. QL selectors are fully serializable, enabling transparent logging and replay of selection operations. A selector is constructed by chaining entry-point functions (e.g., $\mathtt{ql.edges()}$, $\mathtt{ql.faces()}$), predicates (e.g., $\mathtt{ql.tag()}$, $\mathtt{ql.meta()}$), ordering keys (e.g., $\mathtt{ql.center\_axis()}$, $\mathtt{ql.geo()}$), and cardinality constraints (e.g., $\mathtt{.take(n)}$, $\mathtt{.exactly(n)}$). Crucially, QL unifies two complementary selection modes: querying semantic lineage derived from $\mathtt{TopoDelta}$-based tracking, and direct geometric selection based on shape properties. The example below illustrates both behaviors within a unified workflow.

\item 
\begin{minipage}[t]{\linewidth}
\textbf{Example Usage:}
\begin{lstlisting}[language=Python]
from simplecadapi import *

# Base solid with user tags and metadata
base = make_box_rsolid(width=12, height=8, depth=4)
base.add_tag("base_frame")
base.set_metadata("material", "aluminum")

# Add a boss in a local workplane
with SimpleWorkplane(origin=(0, 0, 4)) as _wp:
    boss = make_cylinder_rsolid(radius=2.5, height=2)
model = union_rsolidlist(base, boss)

# Through-hole cut (TopoDelta tags attached)
tool = make_cylinder_rsolid(radius=1.0, height=6)
model = cut_rsolidlist(model, tool)

# Semantic selection via topology tracking
new_cut_faces = (
    ql.faces().where(
        ql.and_(
            ql.tag("origin.tool"),
            ql.meta("track.event", "==", "generated"))
    ).resolve(model)
)
# Geometry-based selection (order-independent)
top_hole_edge = (
    ql.edges().where(ql.curve_type("circle"))
    .order_by(ql.center_axis("z"), desc=True)
    .take(1).exactly(1)
)
# Apply chamfer
result = chamfer_rsolid(model, top_hole_edge, distance=0.5)
\end{lstlisting}
\end{minipage}
\end{itemize}

\begin{table*}[t]
\centering
\caption{Summary of core CAD modeling functions by category}
\label{tab:core_functions}
\begin{tabular*}{\textwidth}{@{\extracolsep{\fill}}lcc@{}}
\toprule
\textbf{Category} & \textbf{Examples} & \textbf{Count} \\
\midrule
Creation & \mcode{make\_box\_rsolid}, \mcode{make\_cylinder\_rsolid}, \mcode{make\_circle\_rwire} & 25 \\
Transformation & \mcode{translate\_shape}, \mcode{rotate\_shape}, \mcode{mirror\_shape} & 3 \\
3D Operations & \mcode{extrude\_rsolid}, \mcode{revolve\_rsolid}, \mcode{loft\_rsolid}, \mcode{sweep\_rsolid} & 4 \\
Boolean & \mcode{union\_rsolidlist}, \mcode{cut\_rsolidlist}, \mcode{intersect\_rsolidlist} & 3 \\
Advanced Features & \mcode{fillet\_rsolid}, \mcode{chamfer\_rsolid}, \mcode{shell\_rsolid}, \mcode{helical\_sweep\_rsolid} & 4 \\
Pattern & \mcode{linear\_pattern\_rsolidlist}, \mcode{radial\_pattern\_rsolidlist} & 2 \\
Tagging \& Selection & \mcode{set\_tag}, \mcode{ql.tag}, \mcode{ql.meta}, \mcode{ql.where}, \mcode{ql.order\_by} & 14 \\
Export & \mcode{export\_step}, \mcode{export\_stl} & 2 \\
\bottomrule
\end{tabular*}
\end{table*}

\subsubsection{API Module}

The API module builds upon the Core module and provides higher-level composite operations and LLM-friendly interfaces for advanced modeling tasks. Its design mainly includes the following two aspects:

\begin{itemize}[leftmargin=*]
    \item \textbf{Core Modeling Functions}: A comprehensive set of over 50 CAD modeling functions organized into categories such as creation, transformation, 3D operations, boolean operations, advanced features, pattern generation, tagging \& selection, and export. These functions allow users or agents to flexibly compose complex modeling procedures while maintaining explicit state transitions and clear semantics. For user-facing convenience, some primitive constructors expose semantic dimension names such as $\mathtt{width}$, $\mathtt{height}$, and $\mathtt{depth}$, but the API documentation binds them explicitly to the world axes as $x$, $y$, and $z$, respectively. A summary of the function categories, representative examples, and counts is shown in Table~\ref{tab:core_functions}.

    \item \textbf{Automation and Self-Maintenance}: To maintain consistency and support continuous evolution, the API module incorporates an automation pipeline with the following functions:
    \begin{enumerate}[label=\textbf{(\arabic*)}]
        \item \textbf{Source Code Parsing}: Automatically parses user-defined Python source files to extract and integrate new or updated modeling functions into the public API.
        \item \textbf{Function Categorization}: Scans and categorizes source code to generate up-to-date API initialization files and alias mappings for consistent public exposure.
        \item \textbf{Documentation Generation}: Extracts function signatures and docstrings to produce structured markdown documentation, improving usability and maintainability.
        \item \textbf{Knowledge Base Synchronization}: Synchronizes generated documentation with an external knowledge base, enabling keyword-based indexing and improved traceability.
    \end{enumerate}
\end{itemize}

\subsection{LLM-Friendly CAD Code Design}

Beyond the ECIP paradigm and its API implementation, we incorporate several complementary design features aimed at improving LLM agents' ability to accurately generate and debug CAD modeling code.

\subsubsection{Structured Error Information}

The error handling mechanism of CadQuery typically produces generic or partially unstructured error messages, which can make automated debugging and code correction more challenging in some scenarios. To facilitate more effective guidance for LLM-driven code generation, ECIP introduces fine-grained structured error messages at the atomic operation level, represented as triples:
\begin{equation}
\mathtt{ErrMsg} = (\mathtt{ErrCau},\ \mathtt{ErrLoc},\ \mathtt{CorrAct}),
\end{equation}
where $\mathtt{ErrCau}$ identifies the root cause of the failure, $\mathtt{ErrLoc}$ specifies the exact location of the error in the script, and $\mathtt{CorrAct}$ suggests potential corrective actions. This structured format provides actionable feedback that LLMs can leverage to improve fault localization and enable automated code repair.

\subsubsection{Explicit Type Annotations}

Type ambiguity commonly leads to erroneous code generation, particularly when dealing with closely related geometric entities such as Wire and Edge. ECIP mitigates this issue by enforcing explicit type annotations through a disciplined naming convention for functions, following the pattern:
\begin{equation}
\mathtt{ActionName\_rReturnType},
\end{equation}
where \(\mathtt{ActionName}\) denotes the CAD operation, \(\mathtt{r}\) is a fixed prefix indicating "returns", and \(\mathtt{ReturnType}\) explicitly declares the type of the returned object. This ensures that even without inspecting documentation, the function signature itself conveys type information, reducing ambiguity during LLM inference. Each function is also accompanied by comprehensive, type-annotated documentation. This explicit type information reduces ambiguity and improves code generation accuracy by LLMs.

\subsubsection{Self-Evolving Composite Operations}

To facilitate knowledge accumulation and abstraction, ECIP supports the definition of \emph{composite operations}--higher-level constructs formed by encapsulating sequences of atomic operations. These composites are saved as new operations within the language, enabling agents to invoke them directly in subsequent tasks.

This self-evolving mechanism effectively expands the language's set of available operations, allowing reuse of complex modeling patterns such as screws, flanges, or other parametric features. In principle, this can improve one-shot generation success by allowing LLM agents to reuse accumulated domain knowledge more efficiently.

\subsection{Knowledge Base for Code Framework}

LLMs possess strong generative and reasoning abilities but often struggle with the fine-grained semantic requirements of CAD APIs, such as precise parameter formats, geometric constraints, and operation-specific return types. These limitations frequently lead to misinterpreted command signatures or hallucinated inputs, resulting in unstable code generation.

To provide explicit, domain-grounded guidance during modeling, we construct a structured knowledge base~$\mathcal{K}$ coupled with retrieval-augmented generation (RAG). Retrieved function annotations and example cases supply accurate API semantics that complement the LLM's implicit priors, helping improve generation reliability.
The knowledge base~$\mathcal{K}$ consists of two key components:

\begin{itemize}[leftmargin=*]
\item \textbf{Function Annotations} ($\mathtt{Anno}$): Semantic descriptions of API functions, including parameter formats, return types, and usage notes;
\item \textbf{Case Examples} ($\mathtt{Case}$): Initially composed of manually selected examples and gradually expanded during use.
\end{itemize}

To automate the construction of $\mathcal{K}$, we design a rule-based pipeline that extracts and organizes information directly from the source code. Specifically, we collect function signatures, parameter types, and return types from Python annotations and type hints, and align them with the corresponding natural language descriptions provided in function-level docstrings. This process yields a structured and semantically consistent API documentation corpus.

\subsubsection{Structured Representation for Retrieval}

We further organize the extracted knowledge into a standardized documentation format to support RAG. Each API entry is represented as a structured Markdown document following a unified template, including function signatures with type annotations, purpose descriptions, parameter specifications, return values, possible exceptions, and practical usage examples. All entries are presented using second-level Markdown headings for structural consistency.

This standardized representation improves retrieval accuracy, reduces ambiguity in natural language queries, and enhances interpretability for both models and developers. For example, a CAD operation such as \(\mathtt{chamfer\_rsolid}\) specifies its input types (e.g., a \(\mathtt{Solid}\) object, a list of \(\mathtt{Edge}\) objects, and a \(\mathtt{float}\) distance parameter) together with usage examples, enabling the retriever to match queries with correct API semantics and usage patterns.

\subsubsection{Semantic-Friendly Chunking}

When building the vectorized representation of $\mathcal{K}$, we employ a customized \textbf{semantic-friendly chunking} strategy based on the standardized Markdown structure of API documents. Specifically, each second-level heading (\(\mathtt{\#\#}\)) is treated as a boundary defining a semantic block. This design ensures \textbf{semantic completeness} (each chunk forms a self-contained unit such as a method or parameter description), \textbf{semantic independence} (different chunks can be processed and retrieved modularly), and \textbf{structural alignment} (the chunking respects the inherent document structure and avoids arbitrary or mid-sentence splits).

Compared to naive fixed-length chunking, this strategy preserves the logical integrity of API documentation and ensures that retrieved content remains coherent and directly usable during generation.

To further improve retrieval precision, we append \textbf{anchor keywords} to each chunk, which explicitly summarize its semantic context. These anchors consist of the API name, the section label, and a small set of representative content tokens. For example, a chunk corresponding to the purpose section of \(\mathtt{chamfer\_rsolid}\) may include:
\[
\mathtt{[chamfer\_rsolid, purpose, chamfer, edge, transition]}
\]

These anchors act as explicit semantic signals that enhance embedding-based retrieval and relevance scoring, improving both recall and precision while preserving the original document semantics. Overall, this design ensures that retrieval at inference time yields coherent and contextually complete knowledge snippets, effectively guiding downstream code generation and improving the robustness of LLM-based CAD reasoning.

\section{Experiments and Comparison}

This section details the implementation of CADDesigner and the comprehensive experimental evaluation conducted to assess its performance in CAD modeling code generation.

\subsection{Implementation Details}
\label{sec:implementation}

We evaluate our CAD modeling code generation algorithm on a workstation with an 8-core Intel CPU under Python 3.12. The system is implemented on a customized agent framework that leverages retrieval-augmented generation (RAG) via RAGFlow. For embedding-based retrieval, we utilize the embedding model bge-large-zh-v1.5 to encode knowledge base documents. The retrieval employs a hybrid similarity scoring method, where vector similarity and keyword-based similarity are combined with weights of 0.6 and 0.4, respectively. The top 3 most relevant documents (top-k=3) are retrieved for downstream generation. For LLM services, the primary agent and the code generation tool (\(T_2\)) employ Claude-4-Sonnet, while requirement refinement (\(T_1\)) and both visual feedback sub-tools (visual question generation and visual feedback generation) are based on Gemini-2.5-Flash. Inference uses a temperature of 0.7 and nucleus sampling (top-p) of 0.9 to balance diversity and coherence. All experiments are conducted without the user feedback loops ($R$ and $Sat$), so the reported results do not include online case accumulation driven by user approval.

\subsection{Dataset}
\label{sec:dataset}
Our experiments are conducted on the large-scale DeepCAD dataset \cite{wu2021deepcad}, which contains approximately 178K parametric CAD models represented in a sketch-and-extrude format. Following SkexGen \cite{xu2022skexgen}, we first remove duplicate models to reduce redundancy. From the training set, we randomly select 1$K$ models to initialize the RAG knowledge base. 
In addition, we develop a conversion script to transform the parametric command sequences of each model into ECIP code snippets, which are included in the RAG knowledge base as $\mathtt{Case}$. For evaluation, we construct two test subsets from the de-duplicated DeepCAD test set. The first subset contains 200 models, selected using uniform stratified sampling based on the number of commands in the sequence (1-10, 11-20, 21-30, 31-40, 41+) to ensure coverage across different complexity levels. The second subset contains 1$K$ models randomly sampled from the same de-duplicated test set and is used for comparison with prior text-to-CAD methods. For each sampled model, a single representative image is rendered from a fixed camera viewpoint to provide visual input for the experiments, while the textual inputs are derived from the abstract-level descriptions provided in the DeepCAD dataset.

\subsection{Evaluation Metrics}
To assess the geometric fidelity of the generated CAD models, we adopt several standard shape similarity metrics. Let $\mathcal{S}$ denote the reference models and $\mathcal{G}$ represent the generated models, with $\mathcal{X}$ and $\mathcal{Y}$ representing their corresponding point clouds. These metrics evaluate the agreement between generated and ground-truth geometries from volumetric and point-wise perspectives.

\textbf{Intersection over Union (IoU)} is utilized to measure the similarity between generated models and the ground truth.
\begin{equation}
    \mathrm{IoU}(\mathcal{G}, \mathcal{S}) = \frac{\mathcal{G} \cap \mathcal{S}}{\mathcal{G} \cup \mathcal{S}},  
\end{equation}
where \(\mathcal{G} \cap \mathcal{S}\) denotes the overlapping volume between the reference and generated models, and \(\mathcal{G} \cup \mathcal{S}\) represents their combined volumetric union. A value of $1$ indicates perfect alignment, while $0$ signifies no overlap.  

\textbf{Chamfer Distance (CD)} calculates point-wise proximity between point clouds sampled from \(\mathcal{X}\) and \(\mathcal{Y}\):  
\begin{equation}
\begin{split}
\mathrm{CD}(\mathcal{X}, \mathcal{Y}) = & \frac{1}{|\mathcal{X}|} \sum_{x \in \mathcal{X}} \min_{y \in \mathcal{Y}} \| x - y \|_2^2 \\
& + \frac{1}{|\mathcal{Y}|} \sum_{y \in \mathcal{Y}} \min_{x \in \mathcal{X}} \| y - x \|_2^2.
\end{split}
\end{equation}

\textbf{Hausdorff Distance (HD)} is a metric for quantifying the similarity between two point sets, primarily used for evaluating the resemblance of object contours.
\begin{equation}
\begin{split}
    \mathrm{HD}(\mathcal{X}, \mathcal{Y}) = \max \big\{& \sup_{x \in \mathcal{X}} \inf_{y \in \mathcal{Y}} d(x, y), \\
    & \sup_{y \in \mathcal{Y}} \inf_{x \in \mathcal{X}} d(x, y)  \big\},
\end{split}
\end{equation}
where \( d(x, y) \) is the Euclidean distance between points \( x \) and \( y \),  \( \sup \) (supremum) and \( \inf \) (infimum) represent the least upper bound and greatest lower bound, respectively.

To ensure fair and reproducible comparison, all geometric metrics are evaluated under standardized \textbf{metric computation protocols}, including normalization, alignment, voxelization, and point cloud sampling, which are applied consistently across all methods and baselines:

\begin{itemize}[leftmargin=*]
  \item \textbf{Normalization and Alignment:} All meshes and sampled point clouds are first normalized to fit within a unit cube \([-0.5,0.5]^3\). For pairwise comparisons, rigid alignment between predicted and ground-truth shapes is performed using the Iterative Closest Point (ICP) algorithm to eliminate differences due to translation and rotation.
  \item \textbf{IoU Voxelization:} For IoU computation, both meshes are voxelized using a fixed voxel size of 0.02. The voxel grid is defined over the union of the bounding boxes of both models, with an additional margin of two voxels to ensure full coverage.
  \item \textbf{Point Cloud Sampling:} For Chamfer Distance (CD) and Hausdorff Distance (HD), 2048 points are uniformly sampled from the surface of each mesh. This sampling density is kept consistent across all methods.
\end{itemize}

In addition, we introduce several process-oriented metrics to analyze modeling stability, reliability, and efficiency under ablation settings:
\textbf{Pass@1} denotes the percentage of cases where valid code is generated on the first attempt, reflecting generation accuracy.
\textbf{AVG Re} (Average Retry) captures the mean number of retries required during the iterative correction loop, indicating convergence efficiency.
\textbf{SUC} denotes the proportion of cases where a syntactically valid CAD model is successfully generated, i.e., models that can be executed and rendered without errors. Note that a syntactically valid model is not necessarily semantically aligned with the input description.
\textbf{Tokens} measures the average number of tokens consumed to generate a single CAD model.
\textbf{Latency} measures the average time required to generate a single CAD model.

\subsection{Comprehensive Ablation Study}
To comprehensively verify the design choices and the effectiveness of different components in our CADDesigner agent, we conduct four groups of ablation studies using the 200-model test subset described in Section~\ref{sec:dataset}: one focusing on the ECIP design, another evaluating the impact of CAD tools and input modalities, a third analyzing the effect of different LLM backends, and a fourth investigating how model complexity influences token usage and generation latency. Unless otherwise specified, all experiments are conducted with text-only input.

\subsubsection{Effect of ECIP Design Components}
To illustrate the benefits of the ECIP design choices, such as the error handling mechanism that is friendly to LLMs and the overall architecture of the ECIP system itself, we evaluate three ECIP variants (full ECIP, ECIP without the detailed error system, and ECIP without the return type annotations).

\begin{table}[!ht]
    \centering
    \caption{Ablation study of ECIP design components on 200 test models.}
    \resizebox{0.42\textwidth}{!}{%
    \begin{tabular}{lccc}
    \toprule
         & \textbf{Pass@1$\uparrow$} & \textbf{AVG Re$\downarrow$} & \textbf{SUC$\uparrow$} \\ \midrule
        ECIP & \textbf{0.45} & \textbf{1.86} & \textbf{100.0\%} \\  
        ECIP w/o Err & 0.41 & 2.62 & 81.5\% \\                 
        ECIP w/o Type & 0.32 & 2.30 & 90.5\% \\                 
    \bottomrule
    \end{tabular}%
    }
    \label{tab:component_ablation}
\end{table}

The results in Table~\ref{tab:component_ablation} show that the full ECIP configuration achieves the best overall performance among all variants. When the error handling mechanism is removed (ECIP w/o Err), the Pass@1 decreases slightly from 0.45 to 0.41, while the AVG Re increases from 1.86 to 2.62, and SUC drops to $81.5\%$. This suggests that although the agent can still generate correct code initially, the lack of detailed feedback slows down convergence during iterative correction and reduces the overall success rate. In contrast, removing type annotations (ECIP w/o Type) causes Pass@1 to drop significantly from 0.45 to 0.32, indicating that the initial code generation is more prone to semantic errors without explicit type guidance. The AVG Re (2.30) is higher than full ECIP, but still lower than ECIP without error handling, and SUC also drops to $90.5\%$. This shows that type annotations help encode critical modeling intent into the prompt, yet some type errors can still be corrected through retries. Overall, we observe a clear division of roles: error handling helps the agent quickly recover from mistakes during iterations, and type annotations improve first-attempt accuracy by reducing semantic ambiguity.

\subsubsection{Effect of CAD Tools and Input Modalities}
To further analyze the individual contributions of the key components in CADDesigner, we evaluate seven variants, labeled A to G, which respectively correspond to: A) removal of function annotations \(\mathtt{Anno}\) from the knowledge base, B) removal of basic model cases \(\mathtt{Case}\), C) disabling the requirement refinement tool \(T_1\), D) disabling the visual feedback tool \(T_4\), E) the full CADDesigner pipeline with text-only input, F) full pipeline with image-only input, and G) full pipeline with combined text and image input. Except for variants F and G, all others use text input exclusively.

\begin{table}[htbp]
    \centering
    \caption{Ablation study of CADDesigner components and input modalities on 200 test models.}
    \resizebox{0.42\textwidth}{!}{%
    \begin{tabular}{lccc}
        \toprule
         & \textbf{IoU$\uparrow$} & \textbf{CD$\downarrow$} & \textbf{HD$\downarrow$} \\
        \midrule
        A (w/o \(\mathtt{Anno}\)) & -- & -- & -- \\
        B (w/o \(\mathtt{Case}\)) & 0.2650 & 0.1350  & 0.4690 \\
        C (w/o $T_1$) & 0.2726 & 0.1251 & 0.5595 \\
        D (w/o $T_4$) & 0.2522 & 0.1192 & 0.4793 \\
        E (Text only) & 0.3041 & 0.1236  & 0.4154 \\
        F (Image only) & 0.3518 & 0.1167  & 0.4262 \\
        G (Text + Image) & \textbf{0.3893} & \textbf{0.0693} & \textbf{0.2553} \\
        \bottomrule
    \end{tabular}%
    }
    \label{tab:extended_ablation}
\end{table}

Table~\ref{tab:extended_ablation} summarizes the quantitative results across multiple evaluation metrics. In Model A, removing semantic API annotations ($\mathtt{Anno}$) prevents the agent from reliably identifying available operations and their usage, leading to failure to generate valid models. This indicates that structured function-level annotations play a critical role in grounding agent-based CAD code generation and guiding convergence toward correct solutions.

Models B, C, and D disable key system components--case examples ($\mathtt{Case}$), requirement refinement ($T_1$), and visual feedback ($T_4$), respectively. Compared to the text-only variant (Model E), all three models show degraded performance in terms of IoU (0.2650, 0.2726, 0.2522 vs. 0.3041), suggesting that each component contributes positively to robust and accurate model generation. Specifically, removing structured case examples and requirement refinement increases semantic ambiguity during early-stage generation, while disabling visual feedback leads to the most significant drop in IoU, highlighting its critical role in iterative refinement and overall structural correctness.

Switching the input modality from text-only (Model E) to image-only (Model F) yields a notable performance improvement, suggesting that abstract textual descriptions alone are often insufficient to fully specify geometric details, whereas images provide richer spatial cues for CAD modeling. Model G, which combines both text and image as input, achieves the best results across all metrics, illustrating that textual information can effectively complement visual context by resolving ambiguities difficult to infer from images alone.

Overall, these results indicate that structured API knowledge, coordinated tool usage, and multimodal input jointly contribute to high-fidelity and robust CAD model generation.

\subsubsection{Effect of LLM Backend}

\begin{table}[htbp]
    \centering
    \caption{Ablation study of different LLM backends on 200 test models for ECIP code generation.}
    \resizebox{0.42\textwidth}{!}{%
    \begin{tabular}{lccc}
        \toprule
        \textbf{LLM Backend} & \textbf{IoU$\uparrow$} & \textbf{CD$\downarrow$} & \textbf{HD$\downarrow$} \\
        \midrule
        Claude-4-Sonnet & \textbf{0.3041} & \textbf{0.1236} & \textbf{0.4154} \\ 
        Gemini-2.5-Pro & 0.2951 & 0.1571 & 0.4902 \\ 
        GPT-5.2-Codex & 0.2914 & 0.1604 & 0.5021 \\
        \bottomrule
    \end{tabular}%
    }
    \label{tab:llm_backend_ablation}
\end{table}

To investigate the influence of different LLM backends on CAD code generation, we conduct an ablation study by replacing the LLM backend used in the code generation tool ($T_2$) of CADDesigner with three representative models: Claude-4-Sonnet, Gemini-2.5-Pro, and GPT-5.2-Codex. Each backend is evaluated under the ECIP code paradigm on 200 test models.

The quantitative results are reported in Table~\ref{tab:llm_backend_ablation}. We observe that CADDesigner achieves consistent performance across the three LLM backends evaluated. Claude-4-Sonnet attains the highest geometric fidelity, reflected by slightly higher IoU and lower CD and HD values. Gemini-2.5-Pro and GPT-5.2-Codex show comparable results, with small differences in CD and HD relative to Claude-4-Sonnet. These observations suggest that, among LLMs of similar capability, CADDesigner produces stable and reliable CAD code. The ECIP representation explicitly manages intermediate modeling states, and the knowledge-constrained generation framework provides structured guidance and reference cases, helping the code generation tool ($T_2$) produce scripts that are structurally correct and semantically meaningful. Overall, these results suggest that CADDesigner maintains broadly consistent CAD code generation across these backends, with moderate variations attributable to the underlying LLM.

\subsubsection{Effect of Model Complexity on Inference Cost}

\begin{figure}[tb]
  \centering
  \includegraphics[width=0.47\textwidth]{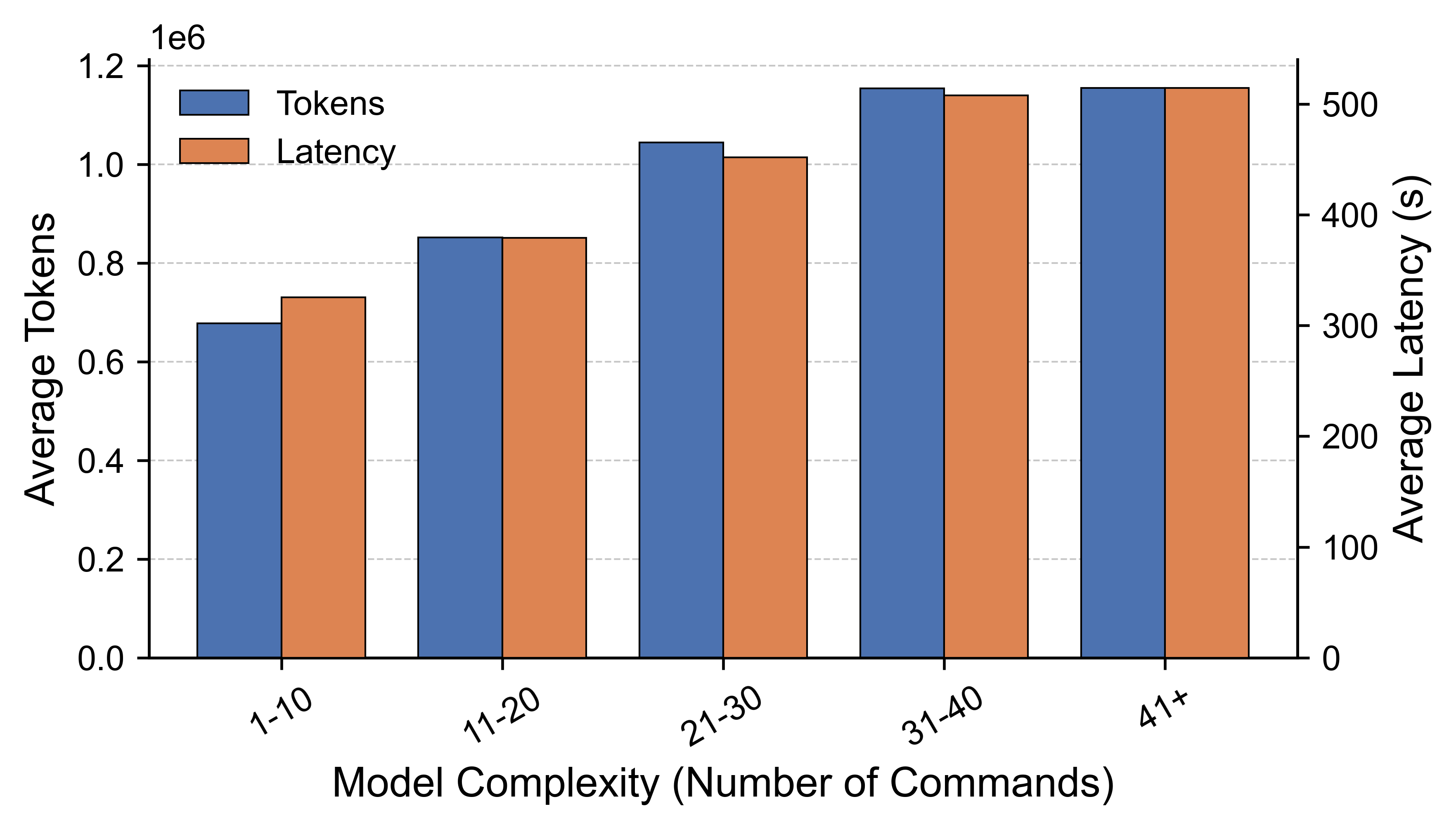}
  \caption{Token usage and generation latency as a function of model complexity (number of commands). Models are grouped into bins of 10 commands each.}
  \label{fig:inference_cost_complexity}
\end{figure}

We evaluate the impact of model complexity on inference cost in CADDesigner. Specifically, we focus on two main metrics: Tokens and Latency. For memory consumption, CADDesigner primarily relies on external LLM APIs, resulting in negligible GPU memory overhead, while the main memory usage comes from RAG components (e.g., RAGFlow), which remain relatively stable and do not scale significantly with model complexity. Models are grouped into bins with an interval of 10 commands. For each complexity bin, we report the average Tokens and average Latency over all models within the bin. As shown in Figure~\ref{fig:inference_cost_complexity}, both Tokens and Latency increase as model complexity grows. This is expected, since more complex models require the agent to generate longer CAD operation sequences and perform more reasoning steps during code generation. However, the increase is relatively moderate rather than steep. From 1--10 to 41 or more commands, token usage and latency grow steadily without abrupt escalation, indicating that the system does not incur excessive overhead as complexity increases. This behavior can be attributed to the design of CADDesigner. The explicit context representation in ECIP and the structured code generation process in \(T_2\) help maintain stable reasoning and avoid redundant or excessively long intermediate steps. As a result, the inference cost scales in a controlled manner with model complexity.

\begin{figure}[tb]
    \centering
    \includegraphics[width=0.47\textwidth]{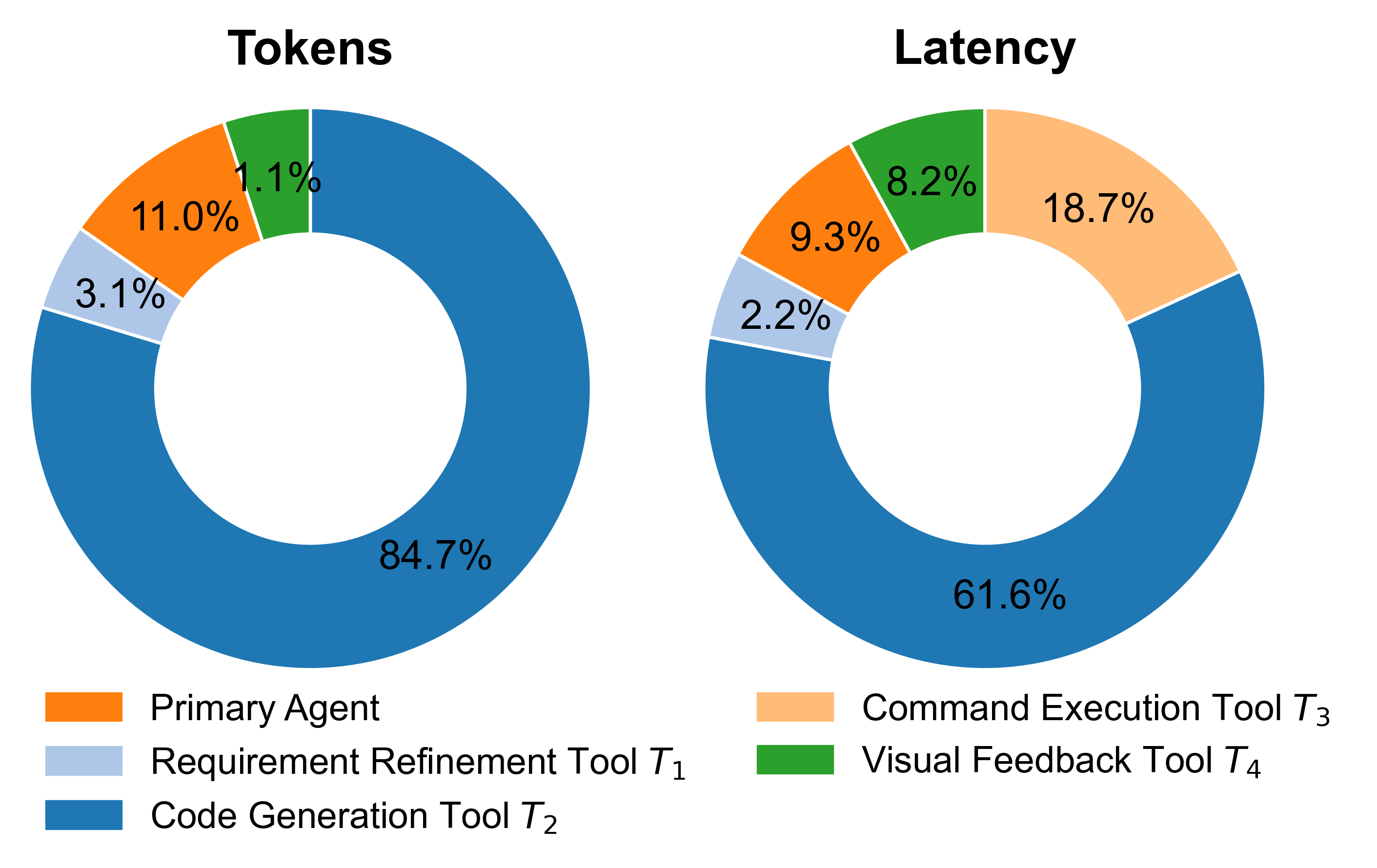}
    \vspace{-10pt}
    \caption{Average inference cost breakdown across the CADDesigner pipeline. Tokens (left) and Latency (right) for each component.}
    \label{fig:combined_tokens_latency_pie}
\end{figure}

We further examine the average contribution of each component in the pipeline across all 200 test models. Figure~\ref{fig:combined_tokens_latency_pie} shows the average proportion of Tokens and Latency for the primary agent and the CAD tools. The code generation tool $T_2$ accounts for the largest share in both Tokens and Latency. The Requirement Refinement Tool $T_1$ and the Visual Feedback Tool $T_4$ each contribute relatively small portions, while the Command Execution Tool $T_3$ contributes modestly to latency without consuming tokens. The primary agent itself has only a minor contribution in both metrics. These observations indicate that $T_2$ is the main driver of both token and latency costs, whereas other components have limited impact on overall inference cost.

\subsection{Comparison Methods}

We compare CADDesigner from two complementary perspectives: (1) different CAD representation paradigms used for code generation, and (2) existing text-to-CAD generation methods.

\subsubsection{Comparison of CAD Representation Paradigms}

\begin{table*}[t]
    \centering
    \caption{Comparison of different CAD representation paradigms on 200 test models.}
    
    \setlength{\tabcolsep}{6pt}
    \renewcommand{\arraystretch}{1.1}
    
    \begin{tabular*}{\linewidth}{@{\extracolsep{\fill}}lcccccccc}
    \toprule
      \textbf{Representation} 
      & \textbf{IoU$\uparrow$} 
      & \textbf{CD$\downarrow$} 
      & \textbf{HD$\downarrow$} 
      & \textbf{Pass@1$\uparrow$} 
      & \textbf{AVG Re$\downarrow$} 
      & \textbf{SUC$\uparrow$}
      & \textbf{Tokens (M)$\downarrow$}
      & \textbf{Latency (s)$\downarrow$} \\
    \midrule
        ECIP & \textbf{0.3041} & \textbf{0.1236} & \textbf{0.4154} & 0.46 & \textbf{1.86} & \textbf{100.0\%} & \textbf{0.98} & 436 \\  
        CadQuery & 0.2827 & 0.1818 & 0.5186 & 0.50 & 2.01 & 87.5\% & 1.40 & 541 \\   
        build123d & 0.2617 & 0.1570 & 0.5126 & \textbf{0.59} & 1.91 & 96.0\% & 1.35 & \textbf{363} \\
    \bottomrule
    \end{tabular*}
    
    \label{tab:representation_comparison}
\end{table*}

We first evaluate the impact of different CAD representation paradigms on code generation performance on 200 test models. Specifically, we compare our proposed ECIP paradigm with two widely used alternatives: CadQuery and build123d. To ensure a fair comparison, we integrate CadQuery and build123d into the same agent framework as CADDesigner by replacing the ECIP-based code generation component, while keeping all other components unchanged as described in Section~\ref{sec:implementation}. We further equip both CadQuery and build123d with RAG modules constructed from their official documentation. All methods use text-only input for evaluation. Here, ECIP, CadQuery, and build123d should be understood as different code representation styles or API interfaces used for generation, rather than fundamentally different geometric kernels.

The quantitative results presented in Table~\ref{tab:representation_comparison} illustrate clear differences among the evaluated CAD code representations. ECIP achieves the best IoU and SUC, as well as the lowest AVG Re, indicating stronger geometric fidelity and more reliable iterative convergence. Since ECIP is implemented as a lightweight representation layer on top of CadQuery, these differences should be understood primarily at the level of code representation and LLM usability rather than the underlying geometric engine. In particular, although CadQuery can also support coding patterns that partially overlap with ECIP, its commonly used fluent interface does not explicitly standardize intermediate modeling context for language models. By contrast, ECIP consistently organizes each operation into explicit object-and-parameter transformations, which makes dependency tracking, code generation, and iterative refinement more reliable for LLM-based agents. Build123d attains the highest Pass@1 and shortest latency, indicating that its compact syntax can often be quickly converted into executable code. However, its lower IoU and higher CD and HD suggest that executable code does not necessarily imply geometric accuracy, which may be due to its reliance on overloaded symbolic operators, making modeling semantics harder for LLMs to interpret than explicit function calls. Overall, the results highlight trade-offs among representation styles for LLM-oriented CAD coding: APIs that simplify LLM code generation can improve early success and reduce execution time, but may compromise geometric accuracy.

\begin{figure*}[!ht]
    \centering
    \includegraphics[width=0.9\textwidth]{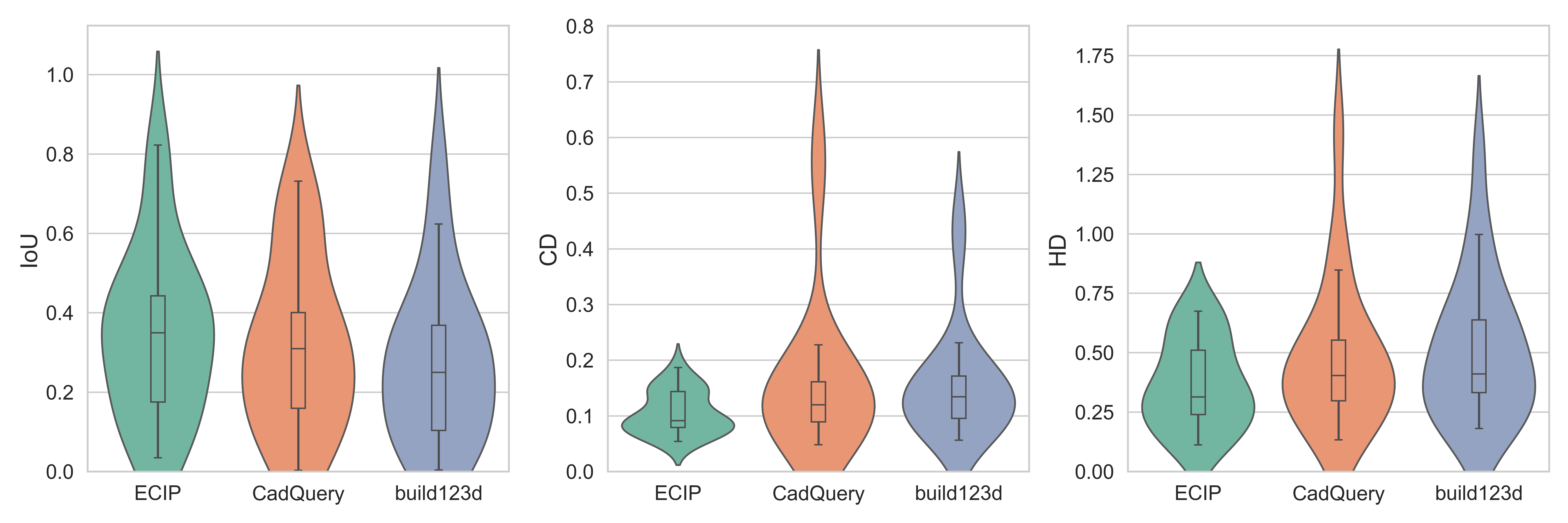}
    \vspace{-10pt}
    \caption{Violin plots comparing the distributions of IoU, CD, and HD across ECIP, CadQuery, and build123d on 200 test models. ECIP achieves higher geometric fidelity and more consistent performance compared to the other paradigms.}
    \label{fig:representation_violin}
\end{figure*}

To further assess robustness and distribution-level differences among ECIP, CadQuery, and build123d, we analyze the distributions of IoU, CD, and HD across the 200 test models. As shown in Figure~\ref{fig:representation_violin}, ECIP exhibits higher median IoU and lower CD/HD values with tighter distributions, indicating more consistent geometric fidelity. In comparison, CadQuery and build123d show wider variability and higher CD/HD, highlighting less precise and less reliable CAD model generation. These results suggest that ECIP demonstrates more accurate and consistent performance across the evaluated models.

\subsubsection{Comparison with Existing Methods}

We further compare our method with two learning-based text-to-CAD generation methods: Text2CAD~\cite{text2cad}, cadrille~\cite{kolodiazhnyi2025cadrille} and one agent-based method: CADCodeVerify~\cite{alrashedy2024generating}. 
\begin{itemize}[leftmargin=*]
    \item \textbf{Text2CAD} proposes an end-to-end transformer-based auto-regressive network to generate parametric CAD models from input texts.
    \item \textbf{cadrille} proposes a two-stage CAD reconstruction pipeline: supervised fine-tuning (SFT) on large-scale procedurally generated data, followed by reinforcement learning fine-tuning using online feedback.
    \item \textbf{CADCodeVerify} proposes an agent that utilizes validation questions and visual feedback for design verification.
\end{itemize}

\begin{table}[htbp]
    \centering
    \caption{Performance comparison of different Text-to-CAD methods under abstract text inputs on 1$K$ test models.}
    \begin{tabular}{lcccc}
        \toprule
        \textbf{Method} & \textbf{IoU$\uparrow$} & \textbf{CD$\downarrow$} & \textbf{HD$\downarrow$} & \textbf{SUC$\uparrow$} \\
        \midrule
        Text2CAD & 0.1831 & 0.1475 & 0.5680 & 96.6\% \\
        cadrille & 0.0274 & 0.2162 & 0.5817 & 98.2\%\\
        CADCodeVerify &   0.2348 &  0.2329  &  0.4892 & 86.1\%  \\
        CADDesigner & \textbf{0.2769} & \textbf{0.1097} & \textbf{0.4347} & \textbf{100.0\%} \\
        \bottomrule
    \end{tabular}
    \label{tab:method_comparison}
\end{table}

\begin{figure}[tb]
  \centering
  \includegraphics[width=0.47\textwidth]{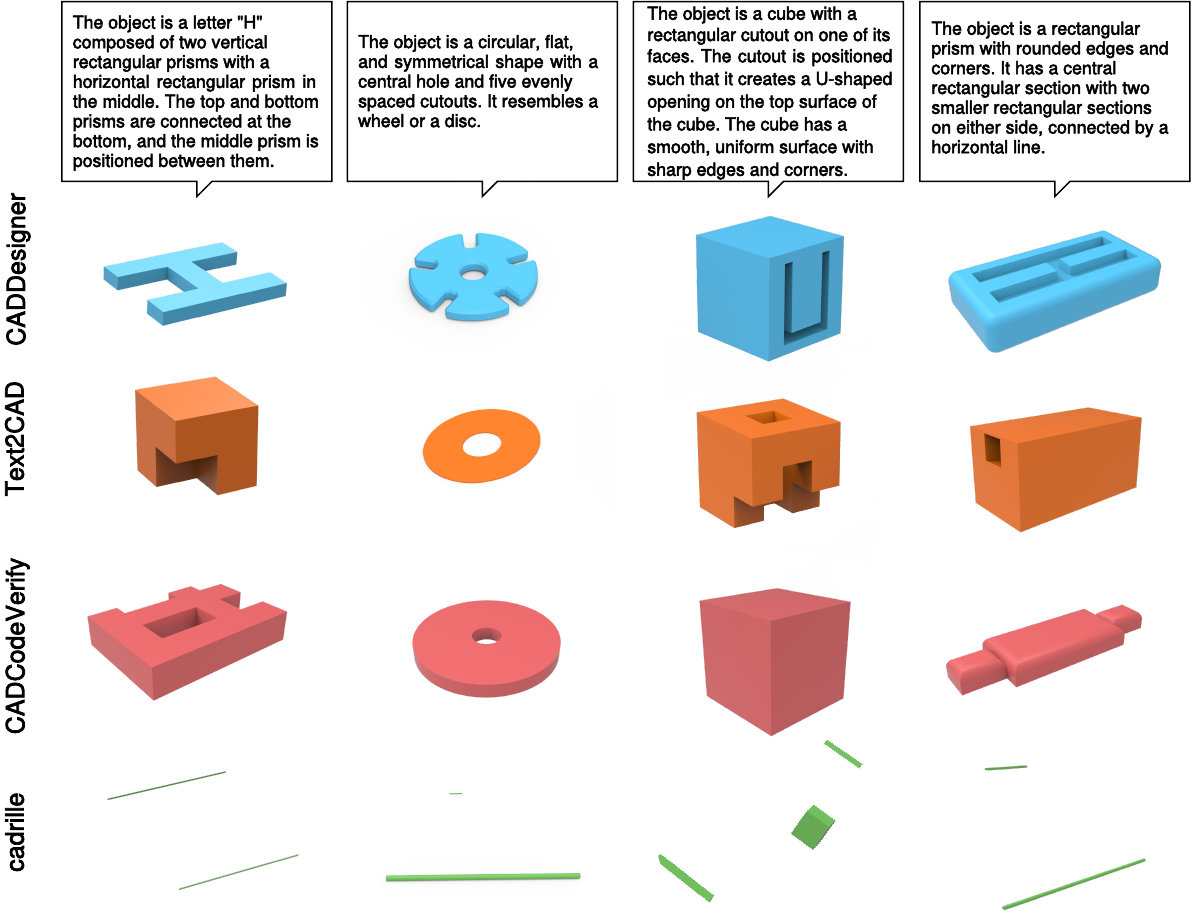}
  \caption{Comparison of generation results across Text2CAD, CADCodeVerify, cadrille, and our method. CADDesigner achieves the best input alignment. Text2CAD and CADCodeVerify show moderate performance, while cadrille generates syntactically valid but semantically incorrect outputs due to poor generalization from expert-level training to abstract inputs.}
  \label{fig:arch}
\end{figure}

\begin{figure}[tb]
  \centering
  \includegraphics[width=0.47\textwidth]{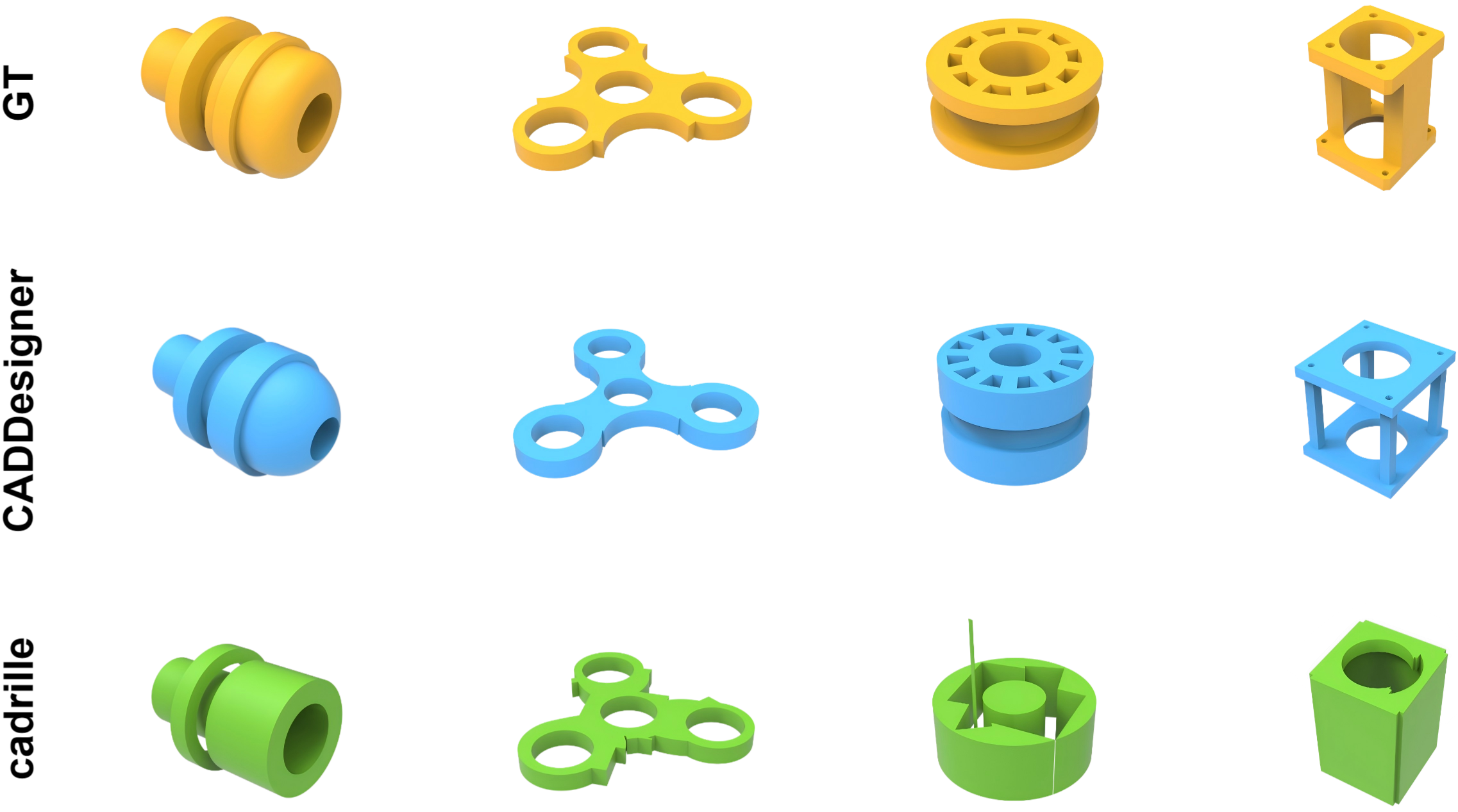}
  \vspace{10pt}
  \caption{Comparison of CADDesigner and cadrille on image-based inputs. CADDesigner (blue) generates CAD models that more closely match the input images on these representative conceptual cases, benefiting from support for operations such as revolve and pattern-based constructions. In contrast, cadrille (green), relying on low-level extrusion, often produces geometrically less faithful results.}
  \label{fig:image_comparison}
\end{figure}

For evaluation, we use the 1$K$-model subset described in Section~\ref{sec:dataset}, randomly sampled from the de-duplicated DeepCAD test set. For each instance, we use the abstract text description as input. We then perform inference using their released weights and our method to generate results for metric calculation. As presented in Table~\ref{tab:method_comparison} and Figure~\ref{fig:arch}, the experimental results show that CADDesigner achieves the best performance. CADDesigner demonstrates the best prompt-result alignment, followed by CADCodeVerify and Text2CAD. While cadrille achieves a high SUC score, indicating it can produce syntactically valid CAD models in most cases, the visualized results reveal these outputs are typically meaningless with respect to the input instructions (see Figure~\ref{fig:arch}). This failure is due to cadrille's poor generalization from expert-level to abstract instructions: it generates structurally valid but semantically irrelevant models. In contrast, the other methods generate outputs that are both syntactically and semantically valid to varying degrees, with our method being the most consistent.

Additionally, we compare our method with cadrille on image-based inputs to further illustrate their respective modeling capabilities. As shown in Figure~\ref{fig:image_comparison}, CADDesigner consistently generates CAD models that closely match the input images, benefiting from a more expressive CAD operation space. In particular, several models require rotationally symmetric structures that are naturally constructed via revolve operations, which are explicitly supported by CADDesigner but not by cadrille, leading to missing or distorted geometries in the latter. Moreover, models with rich geometric details and repetitive structures are more effectively specified using high-level CAD operations such as arrays or patterned constructs. CADDesigner can leverage these higher-level APIs to compactly and precisely generate such structures, whereas cadrille relies primarily on low-level extrusion-based operations, often resulting in geometrically less faithful outputs.

\begin{figure*}[!b]
  \centering
  \includegraphics[width=1.0\textwidth]{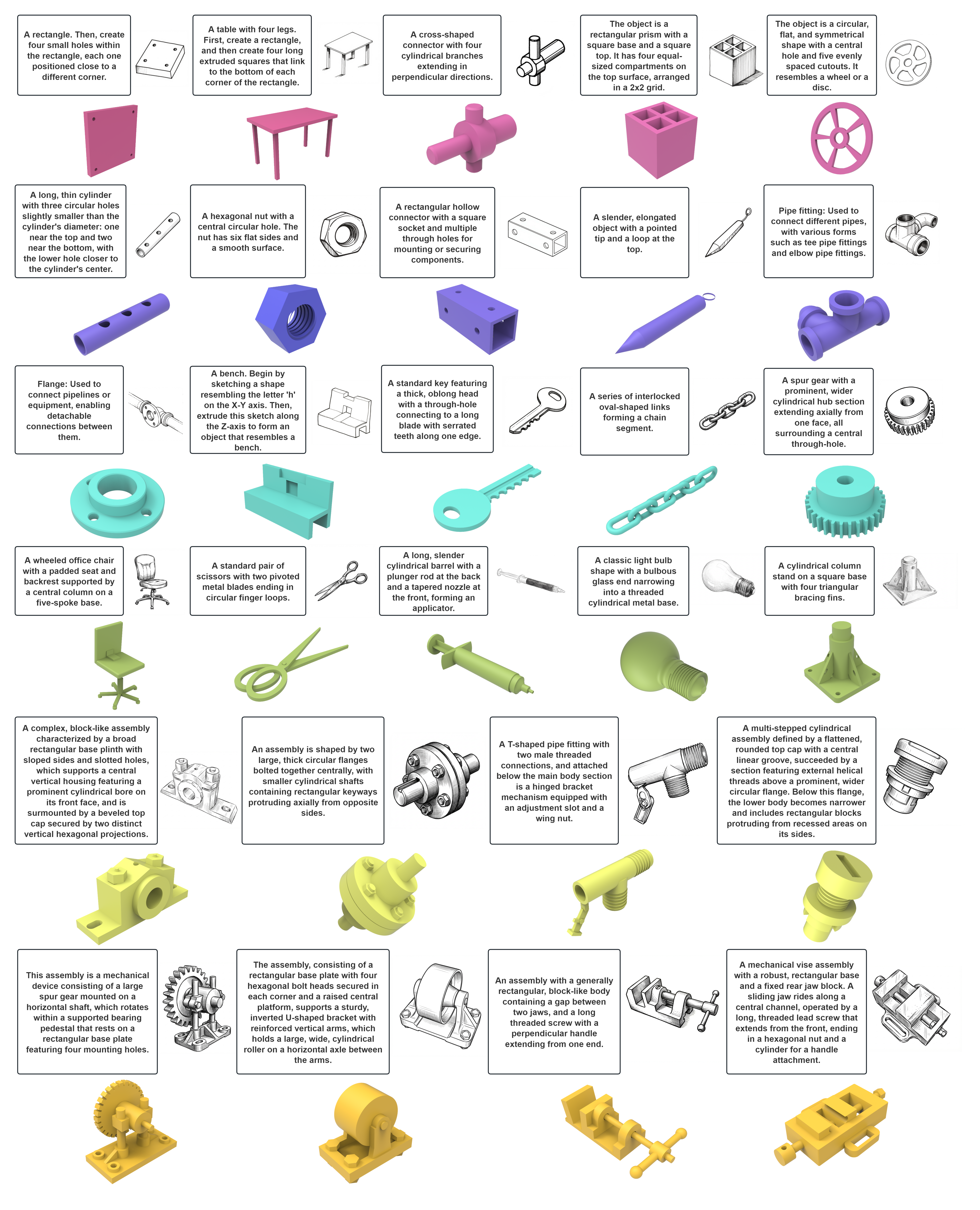} 
  \caption{Visual results with sketch-text input.}
  \label{fig:visual-result}
\end{figure*}

\clearpage

These results highlight that beyond syntactic validity, the availability of expressive and semantically aligned CAD operations is crucial for faithful image-based CAD model generation.

\subsection{Visual Results Presentation}
We additionally provide qualitative results generated from sketch--text inputs, with representative examples illustrated in Figure~\ref{fig:visual-result}. The content presented here is derived from the DeepCAD dataset, the CADCodeVerify dataset, and datasets covering a wide range of geometries, including standard industrial components (e.g., bolts, flanges, and pipe fittings), as well as more complex models and assemblies.

\subsection{Limitations}
Although the latency of CADDesigner scales in a relatively controlled manner with increasing model complexity, its end-to-end response time remains non-negligible in interactive usage scenarios, particularly for complex model generation and multi-round iterative refinement. Our analysis as shown in Figure~\ref{fig:combined_tokens_latency_pie} reveals that the code generation module ($T_2$) is the primary contributor to the overall system latency. Therefore, improving inference efficiency remains a critical challenge.

Furthermore, the current framework does not yet sufficiently incorporate the domain-specific knowledge required in industrial design practice, such as manufacturing constraints, tolerance requirements, assembly semantics, and standardized engineering conventions. Therefore, the capability for complex model generation and validation in specific industrial domains still requires further improvement.

\section{Conclusion and Future Work}
We propose CADDesigner, a novel framework for CAD modeling code generation, combined with an explicit context imperative paradigm to produce high-quality CAD modeling scripts. Experimental results show that CADDesigner achieves competitive performance and outperforms the evaluated baselines in conceptual CAD design tasks. CADDesigner is particularly well-suited for rapid prototyping and early-stage conceptual CAD modeling, where user intent is often abstract and iterative refinement plays a critical role.

In future work, we plan to extend CADDesigner by incorporating learning-based methods that accept point clouds and B-rep data as input, allowing the system to learn geometric and topological constraints through data-driven training. Furthermore, we aim to further optimize the inference efficiency of the framework and introduce domain-specific industrial cases and functional constraints, thereby improving the generation capability and efficiency of our method for complex models in industrial scenarios.

\bibliographystyle{unsrt}


\begin{thebibliography}{10}

\bibitem{wu2021deepcad}
Rundi Wu, Chang Xiao, and Changxi Zheng.
\newblock {DeepCAD}: A deep generative network for computer-aided design models.
\newblock In {\em Proceedings of the IEEE/CVF International Conference on Computer Vision}, pages 6772--6782, 2021.

\bibitem{li2025mamba}
Xueyang Li, Yunzhong Lou, Yu~Song, and Xiangdong Zhou.
\newblock {Mamba-CAD}: State space model for {3D} computer-aided design generative modeling.
\newblock In {\em Proceedings of the AAAI Conference on Artificial Intelligence}, volume~39, pages 5013--5021, 2025.

\bibitem{khan2024cad}
Mohammad~Sadil Khan, Elona Dupont, Sk~Aziz Ali, Kseniya Cherenkova, Anis Kacem, and Djamila Aouada.
\newblock {CAD-SIGNet}: {CAD} language inference from point clouds using layer-wise sketch instance guided attention.
\newblock In {\em Proceedings of the IEEE/CVF Conference on Computer Vision and Pattern Recognition}, pages 4713--4722, 2024.

\bibitem{text2cad}
Mohammad~Sadil Khan, Sankalp Sinha, Talha~Uddin Sheikh, Didier Stricker, Sk~Aziz Ali, and Muhammad~Zeshan Afzal.
\newblock {Text2CAD}: Generating sequential {CAD} designs from beginner-to-expert level text prompts.
\newblock In {\em Advances in Neural Information Processing Systems}, volume~37, pages 7552--7579, 2024.

\bibitem{li2025cad}
Jiahao Li, Weijian Ma, Xueyang Li, Yunzhong Lou, Guichun Zhou, and Xiangdong Zhou.
\newblock {CAD-Llama}: leveraging large language models for computer-aided design parametric {3D} model generation.
\newblock In {\em Proceedings of the Computer Vision and Pattern Recognition Conference}, pages 18563--18573, 2025.

\bibitem{wang2025cad}
Siyu Wang, Cailian Chen, Xinyi Le, Qimin Xu, Lei Xu, Yanzhou Zhang, and Jie Yang.
\newblock {CAD-GPT}: Synthesising {CAD} construction sequence with spatial reasoning-enhanced multimodal {LLMs}.
\newblock In {\em Proceedings of the AAAI Conference on Artificial Intelligence}, volume~39, pages 7880--7888, 2025.

\bibitem{rukhovich2024cad}
Danila Rukhovich, Elona Dupont, Dimitrios Mallis, Kseniya Cherenkova, Anis Kacem, and Djamila Aouada.
\newblock {CAD-Recode}: Reverse engineering {CAD} code from point clouds.
\newblock {\em arXiv preprint arXiv:2412.14042}, 2024.

\bibitem{cadquery}
{CadQuery contributors}.
\newblock {CadQuery}, February 2026.

\bibitem{willis2020fusion}
Karl D.~D. Willis, Yewen Pu, Jieliang Luo, Hang Chu, Tao Du, Joseph~G. Lambourne, Armando Solar-Lezama, and Wojciech Matusik.
\newblock Fusion 360 gallery: A dataset and environment for programmatic {CAD} construction from human design sequences.
\newblock {\em ACM Transactions on Graphics (TOG)}, 40(4), 2021.

\bibitem{xu2022skexgen}
Xiang Xu, Karl~DD Willis, Joseph~G Lambourne, Chin-Yi Cheng, Pradeep~Kumar Jayaraman, and Yasutaka Furukawa.
\newblock {SkexGen}: Autoregressive generation of {CAD} construction sequences with disentangled codebooks.
\newblock In {\em International Conference on Machine Learning}, pages 24698--24724, 2022.

\bibitem{zhang2025diffusion}
Aijia Zhang, Weiqiang Jia, Qiang Zou, Yixiong Feng, Xiaoxiang Wei, and Ye~Zhang.
\newblock {Diffusion-CAD}: Controllable diffusion model for generating computer-aided design models.
\newblock {\em IEEE Transactions on Visualization and Computer Graphics}, 2025.

\bibitem{guo2022complexgen}
Haoxiang Guo, Shilin Liu, Hao Pan, Yang Liu, Xin Tong, and Baining Guo.
\newblock {ComplexGen}: {CAD} reconstruction by {B-rep} chain complex generation.
\newblock {\em ACM Transactions on Graphics (TOG)}, 41(4):1--18, 2022.

\bibitem{ma2024draw}
Weijian Ma, Shuaiqi Chen, Yunzhong Lou, Xueyang Li, and Xiangdong Zhou.
\newblock Draw step by step: reconstructing {CAD} construction sequences from point clouds via multimodal diffusion.
\newblock In {\em Proceedings of the IEEE/CVF Conference on Computer Vision and Pattern Recognition}, pages 27154--27163, 2024.

\bibitem{li2025caddreamer}
Yuan Li, Cheng Lin, Yuan Liu, Xiaoxiao Long, Chenxu Zhang, Ningna Wang, Xin Li, Wenping Wang, and Xiaohu Guo.
\newblock {CADDreamer}: {CAD} object generation from single-view images.
\newblock In {\em Proceedings of the Computer Vision and Pattern Recognition Conference}, pages 21448--21457, 2025.

\bibitem{chen2025cadcrafter}
Cheng Chen, Jiacheng Wei, Tianrun Chen, Chi Zhang, Xiaofeng Yang, Shangzhan Zhang, Bingchen Yang, Chuan-Sheng Foo, Guosheng Lin, Qixing Huang, and Fayao Liu.
\newblock {CADCrafter}: Generating computer-aided design models from unconstrained images.
\newblock In {\em Proceedings of the Computer Vision and Pattern Recognition Conference}, pages 11073--11082, 2025.

\bibitem{wu2023cad}
Sifan Wu, Amir Khasahmadi, Mor Katz, Pradeep~Kumar Jayaraman, Yewen Pu, Karl Willis, and Bang Liu.
\newblock {CAD-LLM}: Large language model for cad generation.
\newblock In {\em Proceedings of the neural information processing systems conference}, 2023.

\bibitem{wu2024cadvlm}
Sifan Wu, Amir~Hosein Khasahmadi, Mor Katz, Pradeep~Kumar Jayaraman, Yewen Pu, Karl Willis, and Bang Liu.
\newblock {CadVLM}: Bridging language and vision in the generation of parametric {CAD} sketches.
\newblock In {\em European Conference on Computer Vision}, pages 368--384, 2024.

\bibitem{xu2024cad}
Jingwei Xu, Zibo Zhao, Chenyu Wang, Wen Liu, Yi~Ma, and Shenghua Gao.
\newblock {CAD-MLLM}: Unifying multimodality-conditioned {CAD} generation with {MLLM}.
\newblock {\em arXiv preprint arXiv:2411.04954}, 2024.

\bibitem{team2024qwen2}
Qwen Team.
\newblock {Qwen2} technical report.
\newblock {\em arXiv preprint arXiv:2407.10671}, 2024.

\bibitem{xie2025text}
Haoyang Xie and Feng Ju.
\newblock {Text-to-{CadQuery}}: {A New Paradigm} for {CAD} generation with scalable large model capabilities.
\newblock {\em arXiv preprint arXiv:2505.06507}, 2025.

\bibitem{mallis2024cad}
Dimitrios Mallis, Ahmet~Serdar Karadeniz, Sebastian Cavada, Danila Rukhovich, Niki Foteinopoulou, Kseniya Cherenkova, Anis Kacem, and Djamila Aouada.
\newblock {CAD-Assistant}: Tool-augmented {VLLMs} as generic {CAD} task solvers.
\newblock {\em arXiv preprint arXiv:2412.13810}, 2024.

\bibitem{freecad}
Juergen Riegel, Werner Mayer, and Yorik van Havre.
\newblock {FreeCAD}: Open-source parametric {3D} {CAD} modeler, 2024.

\bibitem{li2025seek}
Xueyang Li, Jiahao Li, Yu~Song, Yunzhong Lou, and Xiangdong Zhou.
\newblock {Seek-CAD}: A self-refined generative modeling for {3D} parametric {CAD} using local inference via {DeepSeek}.
\newblock {\em arXiv preprint arXiv:2505.17702}, 2025.

\bibitem{yuan20243d}
Zeqing Yuan, Haoxuan Lan, Qiang Zou, and Junbo Zhao.
\newblock {3D-PreMise}: Can large language models generate {3D} shapes with sharp features and parametric control?
\newblock {\em arXiv preprint arXiv:2401.06437}, 2024.

\bibitem{alrashedy2024generating}
Kamel Alrashedy, Pradyumna Tambwekar, Zulfiqar Zaidi, Megan Langwasser, Wei Xu, and Matthew Gombolay.
\newblock Generating {CAD} code with vision-language models for 3d designs.
\newblock {\em arXiv preprint arXiv:2410.05340}, 2024.

\bibitem{build123d}
Roger Maitland, jdegenstein, Bernhard, Ethan Rooke, JR~Mobley, snoyer, Jojain, Andreas~Felix H{\"a}berle, Ruud, Ami Fischman, Jason~S. McMullan, Roman Dvo{\v r}{\'a}k, simon klemenc, BogdanTheGeek, Spectre5, Dalibor Fr\'{i}valdsk\'{y}, Daniele D'Orazio, George, hoijui, OpenVMP, Yeicor, Alexander Steppke, mayhem 64, luzpaz, nobkd, Victor Poughon, slobberingant, Arno Bosch, Barnaby Walters, and Matti Eiden.
\newblock gumyr/build123d: v0.9.1, feb 2025.

\bibitem{yao2023react}
Shunyu Yao, Jeffrey Zhao, Dian Yu, Nan Du, Izhak Shafran, Karthik~R Narasimhan, and Yuan Cao.
\newblock {ReAct}: Synergizing reasoning and acting in language models.
\newblock In {\em The Eleventh International Conference on Learning Representations}, 2023.

\bibitem{kolodiazhnyi2025cadrille}
Maksim Kolodiazhnyi, Denis Tarasov, Dmitrii Zhemchuzhnikov, Alexander Nikulin, Ilya Zisman, Anna Vorontsova, Anton Konushin, Vladislav Kurenkov, and Danila Rukhovich.
\newblock {cadrille}: Multi-modal {CAD} reconstruction with online reinforcement learning.
\newblock {\em arXiv preprint arXiv:2505.22914}, 2025.

\end{thebibliography}

\end{document}